%% file: paper.tex
\documentclass[a4paper]{article}

\input{packages}

\input{commands}

\input{title}

\newtheorem{theorem}{Theorem}
\newtheorem{prop}{Proposition}
\newtheorem{lem}{Lemma}

\begin{document}

\maketitle

\input{00-Abstract}

\vspace{-5pt}

\section{Introduction}
\label{sec:introduction}
\input{01-Intro}

\section{Related Work}
\label{sec:related}
\input{02-Related}

\section{Notations and Problem Formulation}
\label{sec:problem-formulation}
\input{03-ProblemFormulation}

\section{Approach}
\label{sec:approach}
\input{04-Approach}

\section{Results}
\label{sec:results}
\input{05-Results}

\vspace{-5pt}
\section{Conclusions}
\label{sec:conclusions}
\input{06-Conclusions}


\begin{appendices}
	\input{07-Appendix}
\end{appendices}


\bibliographystyle{plain}


\end{document}

%% file: packages.tex
\usepackage[english]{babel}
\usepackage[utf8]{inputenc}

\usepackage[dvipsnames]{xcolor}
\usepackage{color,xcolor,ucs}
\usepackage{floatrow}
\usepackage{tabularx}
\usepackage{float}
\usepackage{amsfonts}
\usepackage{amsmath}
\usepackage{amssymb}
\usepackage{makeidx}         
\usepackage{comment}         
\usepackage{graphicx}        
\usepackage{multicol}        
\usepackage[bottom]{footmisc}
\usepackage{bm}

\usepackage{cite}
\usepackage{url}

\usepackage{xr-hyper}

\usepackage{algorithm}
\usepackage{algpseudocode}

\usepackage{xspace}
\usepackage{rotating}

\usepackage{graphicx}
\usepackage{threeparttable}
\usepackage{multirow}
\usepackage[font=scriptsize,labelfont=bf]{caption}

\usepackage{subcaption}
\usepackage{wrapfig}

\usepackage[title]{appendix}


%% file: commands.tex
\newcommand{\bydef}{\triangleq}

\newcommand{\prob}[1]{\ensuremath{\mathbb{P}({#1})}}

\algrenewcommand\algorithmicrequire{\textbf{Input:}}
\algrenewcommand\algorithmicensure{\textbf{Input:}}
\algnewcommand{\LineComment}[1]{\State \(\triangleright\) #1}

\newcommand{\brac}[1]{\ensuremath{\left[ {#1} \right]}}
\newcommand{\augplus}{\ensuremath{\boxplus}}
\newcommand{\X}{\ensuremath{X}}

\newcommand{\his}{\ensuremath{h}}

\newcommand{\Z}{\ensuremath{Z}}

\newcommand{\ZSp}{\ensuremath{\mathcal{Z}}}
\newcommand{\Xinv}{\ensuremath{\X^{in}}}

\newcommand{\Xinvp}{\ensuremath{\X^{in+}}}

\newcommand{\Xinvi}[1]{\ensuremath{\X^{in({#1})}}}
\newcommand{\XinvSp}{\ensuremath{\mathcal{\X}^{in}}}
\newcommand{\Xnot}{\ensuremath{\X^{\neg in}}}

\newcommand{\Xnew}{\ensuremath{\X_{new}}}
\newcommand{\XnewSp}{\ensuremath{\mathcal{\X}_{new}}}
\newcommand{\z}{\ensuremath{z}}
\newcommand{\zz}[1]{\ensuremath{z^{({#1})}}}
\newcommand{\x}{\ensuremath{x}}
\newcommand{\xinv}[1]{\ensuremath{\x^{in({#1})}}}
\newcommand{\xnew}[1]{\ensuremath{\x_{new}^{({#1})}}}
\newcommand{\ent}[1]{\ensuremath{\mathcal{H} \left[ {#1} \right]}}
\newcommand{\expec}[1]{\ensuremath{\underset{{#1}}{\mathbb{E}}}}
\newcommand{\ig}[2]{\ensuremath{IG \left[ {#1} ; {#2} \right]}}
\newcommand{\igaug}[3]{\ensuremath{IG_{aug} \left[ {#1} \augplus {#2} ; {#3} \right]}}
\newcommand{\mi}[2]{\ensuremath{I \left[ {#1} ; {#2} \right]}}
\newcommand{\micond}[3]{\ensuremath{I \left[ {#1} ; {#2} \mid {#3} \right]}}
\newcommand{\miaug}[3]{\ensuremath{I_{aug} \left[ {#1} \augplus {#2} ; {#3} \right]}}
\newcommand{\igseq}{\ensuremath{IG_0^t}}
\newcommand{\igseqinv}{\ensuremath{IG{_0^t}^{in}}}
\newcommand{\miseq}{\ensuremath{I_0^t}}
\newcommand{\miseqinv}{\ensuremath{I{_0^t}^{in}}}
\newcommand{\micon}[1]{\ensuremath{I_{{#1}-1}^{#1}}}
\newcommand{\miconinv}[1]{\ensuremath{I{_{{#1}-1}^{#1}}^{in}}}
\newcommand{\FT}{\ensuremath{\mathcal{P}_T}}
\newcommand{\FZ}{\ensuremath{\mathcal{P}_Z}}
\newcommand{\bel}[1]{\ensuremath{b \brac{{#1}}}}

\algrenewcommand\algorithmicrequire{\textbf{Input:}}
\algrenewcommand\algorithmicensure{\textbf{Output:}}

\newcommand{\invmi}{\texttt{involve-MI}}

\newcommand{\mismc}{\texttt{MI-SMC}}
\newcommand{\invmismc}{\texttt{involve-MI-SMC}}

\newcommand{\invmikde}{\texttt{involve-MI-KDE}}

\newcommand{\nkde}{\texttt{Naive KDE}}

%% file: title.tex
\title{\LARGE \bf \invmi{}: Informative Planning with High-Dimensional Non-Parametric Beliefs\thanks{This work was partially supported by the Israel Science Foundation (ISF) and by US NSF/US-Israel BSF.}}

\author{Gilad Rotman$^{1}$ and Vadim Indelman$^{2}$ \\
	$^1$Technion Autonomous Systems Program (TASP) \\
	$^2$Department of Aerospace Engineering\\
	Technion - Israel Institute of Technology, Haifa 32000, Israel\\
	\small{\texttt{gilad.rotman@technion.ac.il, vadim.indelman@technion.ac.il}}}

\date{}


%% file: 00-Abstract.tex
\begin{abstract}

One of the most complex tasks of decision making and planning is to gather information. This task becomes even more complex when the state is high-dimensional and its belief cannot be expressed with a parametric distribution. Although the state is high-dimensional, in many problems only a small fraction of it might be involved in transitioning the state and generating observations. We exploit this fact to calculate an information-theoretic expected reward, mutual information (MI), over a much lower-dimensional subset of the state, to improve efficiency and without sacrificing accuracy. A similar approach was used in previous works, yet specifically for Gaussian distributions, and we here extend it for general distributions. Moreover, we apply the dimensionality reduction for cases in which the new states are augmented to the previous, yet again without sacrificing accuracy.
We then continue by developing an estimator for the MI which works in a Sequential Monte Carlo (SMC) manner, and avoids the reconstruction of future belief's surfaces. Finally, we show how this work is applied to the informative planning optimization problem.
This work is then evaluated in a simulation of an active SLAM problem, where the improvement in both accuracy and timing is demonstrated.


\end{abstract}

%% file: 01-Intro.tex
Planning under uncertainty is of most importance for many applications. Our world is stochastic in nature, thus for every inference and planning task this stochasticity needs to be taken into account, or catastrophes may occur.

Addressing stochasticity can be done in many levels. At the simplest level, the stochasticity is only being considered at the inference phase, while at planning the state is treated as if it is completely deterministic. In one of the next levels, stochasticity is also being considered within planning, however the planner is considered to reach the goal without caring about the uncertainty of the state. In one of the top levels, the uncertainty of the state also needs to be measured. For example, if the plan is to reach a goal state with some minimum probability or if the plan is to purely gain as much information as possible about the state. Such an approach, where we select a sequence of actions based also (or only) upon the certainty of the state, is known as informative planning. For instance, a task which is considered as informative planning is search and rescue, where in the "search" phase an exploration of unknown terrains might be done.

There are a few measures for the uncertainty of a state, which we will refer to as information-theoretic costs or rewards. A very common such cost is entropy. Yet, for many applications we wouldn't be interested in the absolute value of the uncertainty, but rather by how much we expect it to be reduced, or in other words how much information is to be gained. Such a reward is called Information Gain (IG). Also, since the state is not always directly observable,
observations are used. But when evaluating the future, observations themselves are also unknown, thus the uncertainty of these should also be taken into account. 
Consequently, we consider Mutual Information (MI), which is the expected IG over the observations.
We note that the reader might find that some parts of the literature refer to MI also as IG, but as in the artificial intelligence literature, we distinct between the two terms in this paper.

\begin{figure}[t]
	\begin{subfigure}[b]{0.48\textwidth}
		\centering
		\includegraphics[width=0.8\columnwidth]{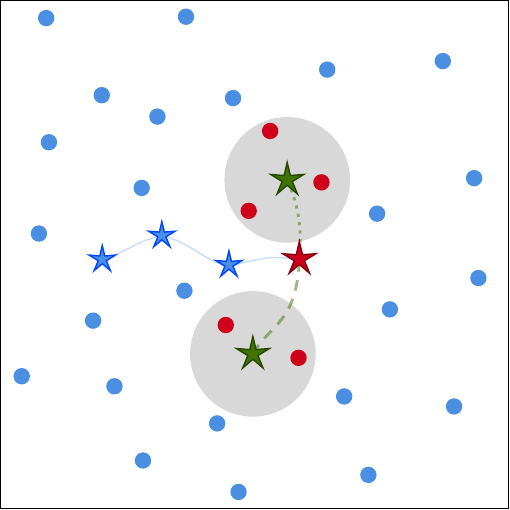}
		\caption{}
		\label{fig:slam_involved}
	\end{subfigure}
	\begin{subfigure}[b]{0.48\textwidth}
		\centering
		\includegraphics[width=0.9\columnwidth]{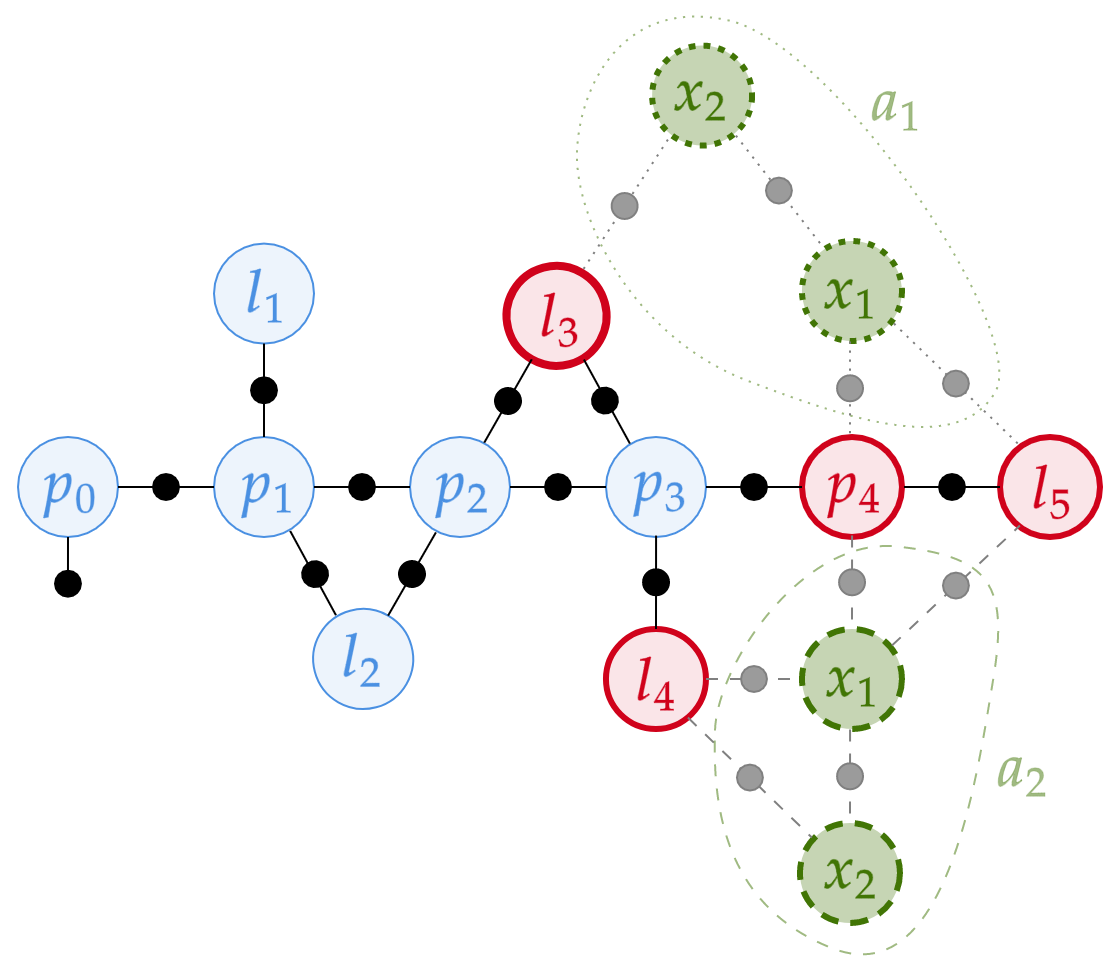}
		\caption{}
		\label{fig:factorgraph_involved}
	\end{subfigure}
	
	\caption{A toy example of an active SLAM problem, where a drone has to choose between two actions. \textbf{(a)} As a 2D map: stars represent drone's poses; dots represent  landmarks; {\color{gray} big circles} are drone's eye of sight from future poses. \textbf{(b)} As a factor graph: circles represent variable nodes; dots represent factors (probabilistic constraints); the unity of all  solid-circled variable nodes represent the prior state $X_0$. In both figures,  colors partition between {\color{red} involved}, {\color{RoyalBlue} uninvolved} and {\color{Green} new (future)} variables. \vspace{-15pt}}
	\label{fig:illustration_involved}
\end{figure}

Naturally, by increasing the dimension of the problem, the complexity of evaluating information-theoretic terms is increased as well, and this relation is exponential (known as the Curse of Dimensionality). To the best of our knowledge, there are no existing approaches which deal with the painful and critical issue of calculating information-theoretic terms in general $\rho$-POMDPs with high-dimensional non-parametric beliefs.

This paper proposes a novel approach to fill this gap. By exploiting structure inherent in many planning problems, it breaks the relation between the dimension of the problem and the complexity of calculating the information-theoretic terms.
Fig.~\ref{fig:slam_involved} shows a toy example of a 2D active SLAM problem to illustrate the structure our approach exploits. An aerial drone observes landmarks from above with a certain sensing range. At current time it needs to choose between two actions. While the state is high-dimensional because it contains many landmarks, we observe that only a few landmarks might be observed at future time, and we will denote these landmarks as the \emph{involved} variables.
The key idea of our approach is to discard the uninvolved landmarks at planning time for the calculation of the expected reward, MI.
We prove that by discarding these variables, the solution is still exact, yet the dimension (and thus the complexity) of the problem is reduced dramatically
%
Furthermore, a naive approach entails also explicit estimation of future beliefs' surfaces, which adds yet another level of errors. Another key contribution of our approach is that it allows skipping this step by having access to the problem's models.
We finalize by showing how these contributions can be used together with existing belief tree solvers.



%% file: 02-Related.tex

The notion that in some cases the correlations between some of the state variables could be discarded for the calculation of entropy, thus improving the efficiency, was first introduced in \cite{Indelman16ral}. This idea was extended to a more general case in \cite{Elimelech17icra} and \cite{Kopitkov17ijrr}. However, these papers addressed the problem assuming Gaussian distributions. The current paper extends these works to the non-parametric case.

Reviewing state-of-the-art approaches for planning with non-parametric beliefs, most do not attempt to address the problem of evaluating the uncertainty of a state (e.g. \cite{Kurniawati08rss}, \cite{Silver10nips}, \cite{Ye17jair}, \cite{Kurniawati16chapter}).
These works consider the POMDP framework, which does not support belief-dependent rewards. The recent approaches presented in e.g. \cite{Sunberg18icaps} and \cite{Fischer20icml},
do support belief-dependent rewards, by using the more general $\rho$-POMDP framework.
Yet, while these are concerned with improving the search over a tree, the calculation of the information-theoretic costs when the state is high-dimensional remains a problem.

Many estimators exist for information-theoretic rewards. A brief overview of the most common estimators for entropy can be found in \cite{Beirlant97jmss}.
Another approach to estimate entropy, a particle-filter based approach, is presented in \cite{Boers10fusion}.
Furthermore, \cite{Ryan08gnc} suggests to use a piece-wise linear approximation of the beliefs' surface.
Yet, all will face the Curse of Dimensionality. The work presented in \cite{Stowell09spl} claims to support high-dimensional spaces, however the simulations show this approach isn't superior for any arbitrary belief compared to other approaches.

When changing the context a bit, and looking at an expected reward, mutual information, \cite{Chli09thesis} and \cite{Zhang20ijrr} both state that the MI over a multi-dimensional variable can be calculated over a lower dimensional subset of this variable. Yet their statements are applied specifically for the problems of feature matching and active mapping, respectively. 
Also, these do not address problems in which quantifying the information should also take into account the state might change.

To the best of our knowledge, there are two works which, at some level, construct planners with an information-gathering task and which support high-dimensional non-parametric beliefs with complexity not exponential with the dimension of the state, yet do not support the more general setting we are addressing in this paper. One of which is \cite{Stachniss05rss}, addressing the very specific active SLAM problem. It exploits the unique structure of SLAM to use Rao-Blackwellization, which might not be useful for more general problems. Also, it approximates the joint entropy by averaging the individual entropies of the state variables, thus breaks up the correlations between them.

The second work is \cite{Platt11isrr}, which uses a slightly different and more specific formulation than the general POMDP, where the objective is to minimize the expected cost. In its formulation, the objective is to reach a desired region of the state space with a guaranteed minimum probability of success. This, in turn, means that it wouldn't necessarily choose the most informative path. Also, it does not support, at least directly, information-theoretic rewards. And, lastly, this approach was eventually tested only on problems with up to four dimensions.

Our approach deals with the limitations of the former.



%% file: 03-ProblemFormulation.tex
\subsection{$\rho$-POMDPs}

\sloppy 
Facing the task of informative planning, we will use a model which is an extension of the well-known Partially Observable Markov Decision Process (POMDP). This extension is referred to as $\rho$-POMDP and is modeled as a tuple $\langle \mathcal{X}, \mathcal{A}, \mathcal{Z}, \bel{\X_0}, \mathbb{P}_T, \mathbb{P}_Z, \rho \rangle$, where $\mathcal{X}$ is the state space, $\mathcal{A}$ is the action space, and $\mathcal{Z}$ is the observation space; $\bel{\X_0} \bydef \prob{\X_0}$ is the prior belief over the state,  $\mathbb{P}_T$ is the probabilistic transition model, and $\mathbb{P}_Z$ is the probabilistic observation model; $\rho \left( \bel{\X_t}, a_t \right)$ is a belief-dependent immediate reward function.


The prior belief and probabilistic models are used in the inference layer for determining future posterior beliefs. The belief over the state $\X_t$ at time $t$ is defined by $\bel{\X_t} \bydef \prob{\X_t \mid \his_t}$,
where $\his_t = \left\{ a_{0:t-1}, \z_{1:t} \right\}$ is the history, containing all actions $a_{0:t-1}$ and observations $\z_{1:t}$ acquired up to time $t$.
The transition model $\mathbb{P}_T \bydef \prob{x_t \mid \X_{t-1}^{tr}, a_{t-1}}$ defines the distribution of the successor state $x_t$, given a subset of its previous state $\X_{t-1}^{tr} \subseteq \X_{t-1}$ participating in the transition, and the chosen action $a_{t-1}$.
In this paper, we use the \emph{smoothing} formulation, where the successor state is \emph{augmented} to the previous, thus constructing the \emph{joint} successor state $\X_t = \left\{ \X_{t-1}, x_t \right\}$. This formulation means that the state's dimension increases in time. 
The observation model $\mathbb{P}_Z \bydef \prob{\z_t \mid \X_t^{obs}}$ defines the conditional distribution of receiving an observation $\z_t$, given a subset of the state $\X_t^{obs} \subseteq \X_t$ which participates in generating the observation.

For example, in the case of active (full) SLAM, the state is defined as the union of all poses and landmarks. The transition model can be formulated between two consecutive poses, and the observation model can be formulated such that an observation is generated given the last pose and a specific landmark. This example is illustrated in the factor graph in Figure~\ref{fig:factorgraph_involved}, where $p$ refers to poses, $l$ to landmarks and the factor nodes represent the probabilistic constraints between the variables, given with the motion and observation models.

Using Bayes' rule and the chain rule, the belief can be defined recursively as $\bel{\X_t} = \eta \FT \FZ \bel{\X_l}$,
where $\FT \triangleq \prod_{i = l+1}^{t} \prob{x_i \mid \X_{i-1}^{tr}, a_{i-1}}$ is the sequential transition model, $\FZ \triangleq \prod_{j = l+1}^{t} \prob{\z_j \mid \X_j^{obs}}$ the sequential observation model, and $\eta^{-1} \triangleq \prod_{k = l+1}^{t} \prob{\z_k \mid \his_k^{-}}$ the sequential normalizer, where $\his_k^{-} \bydef \his_k \setminus \z_k$. 

%
The tasks of the planning scheme are represented with the immediate reward function $\rho \left( \bel{\X_t}, a_t \right)$. Obviously, each planning task might involve multiple different (and sometimes contradicting) tasks, such as energy consumption and time to reach a goal, thus the reward functions might be shaped with multiple different terms. Having the reward defined as belief-dependent allows to express information-gathering tasks. 


Planning $T$ steps into the future, the objective is then to find an action sequence $a_{0:T-1}$ which maximizes the expected sum of rewards, denoted as the objective function, $J \left( \bel{\X_0}, a_{0:T-1} \right) = \underset{\mathcal{Z}_{1:T}}{\mathbb{E}} \left[ \sum_{t=0}^{T-1} \rho_t + \rho_T \right]$,  where $\mathcal{Z}_{1:T}$ is the sequential space of future observations, $\rho_t \bydef \rho \left( \bel{\X_t, a_t} \right)$ is the reward at each time $t$, and $\rho_T \bydef \rho \left( \bel{\X_T} \right)$ is a terminal reward. Due to commutativity, the objective function can also be written as the sum of expected rewards, $J \left( \bel{\X_0}, a_{0:T-1} \right) = \sum_{t=0}^{T-1} \left[ \underset{\mathcal{Z}_{1:t}}{\mathbb{E}} \left[ \rho_t \right] \right] + \underset{\mathcal{Z}_{1:T}}{\mathbb{E}} \left[ \rho_T \right]$, which means that we can evaluate the expected rewards rather than the rewards themselves.

We note that although the problem is formulated as an open loop, our approach also supports a close loop formulation, in which the objective is to find a policy $\pi$ instead of an action sequence. This is since our approach focuses on evaluating the (expected) rewards in the objective function. Focusing on the objective function evaluation also means that other building blocks of the planning task, such as the inference engine and the optimization solver, can be chosen independently.

\subsection{Information-theoretic rewards}

A commonly used information-theoretic reward is negative (differential) entropy. The entropy of the state $\X_t \in \mathcal{X}_t$, distributed with $\bel{\X_t}$, is defined as
\begin{equation}
	\ent{\X_t \mid \his_t} \bydef -\int_{\mathcal{X}_t} \bel{\X_t} \log \bel{\X_t} d\X_t.
\end{equation}
Another commonly used information-theoretic reward is information gain (IG), which quantifies the amount of information gained for a certain variable by knowing the value of another variable. For the case where the state changes between time steps, as presented in the previous subsection, the original definition of IG is insufficient, since it does not account for the additional uncertainty obtained by changes in the state. Thus, we define the reward more generally as the difference between the entropies of the prior state $\X_0$ and the successor state $\X_t$. We remind that in our formulation the posterior state is augmented, i.e. $\X_t = \left\{ \X_0, \x_{1:t} \right\}$, thus we will refer to this reward as \emph{augmented} IG, defined at each time $t$ as
\begin{equation}
	\label{eq:augmented_ig_full}
	\igaug{\X_0}{x_{1:t}}{\Z_{1:t}=z_{1:t} \mid a_{0:t-1}}
	\bydef \ent{\X_0} - \ent{\X_t \mid \Z_{1:t}=z_{1:t},  a_{0:t-1}},
\end{equation}
where $\Z_{1:t}$ represents the observation sequence as a random variable, and the symbol $\augplus$ provides a distinction between the prior state $\X_0$ and the new, augmented subset $x_{1:t}$ of the successor state $\X_t$. This distinction is necessary since $x_{1:t}$ only appears in the posterior entropy term, while $\X_0$ appears in both terms. We then define the corresponding expected reward, \emph{augmented} mutual information (MI), as
\begin{equation}
	\label{eq:augmented_mi_full}
	\begin{aligned}
		\miaug{\X_0}{x_{1:t}}{\Z_{1:t} \mid a_{0:t-1}} 
		&\bydef \underset{\mathcal{Z}_{1:t}}{\mathbb{E}} \Big[ \igaug{\X_0}{x_{1:t}}{\Z_{1:t}=z_{1:t}, a_{0:t-1}} \Big] \\
		&= \ent{\X_0} - \ent{\X_t \mid \Z_{1:t}, a_{0:t-1}}.
	\end{aligned}
\end{equation}
We note that augmented IG and augmented MI are generalizations of the original IG and MI. The definitions of the latter are provided in Appendix \ref{sec:Info_theo}. 

For the purposes of planning, since negative entropy and augmented IG differ only by the value $\ent{\X_0}$, which is constant for each action, using any of these as rewards is equivalent. We further continue to present our approach by choosing augmented IG as the reward, $\rho_t = \igaug{\X_0}{x_{1:t}}{\Z_{1:t}=z_{1:t} \mid a_{0:t-1}}$ (for the terminal reward as well), which dictates that the expected reward is augmented MI. The basis of our approach focuses on the evaluation of the augmented MI, thus we will present it over the augmented MI at time $t$ alone. Naturally, it will apply for the entire horizon. 

For readability, we  denote the prior state as $\X = \X_0$, the state at time $t$ as $\X' = \X_t = \left\{\X_0, x_{1:t} \right\}$ and the augmented part of the state as $\Xnew = \left\{ x_{1:t} \right\}$. Also, the future observation sequence up to time $t$ is denoted as $\Z = \left\{ \Z_{1:t} \right\}$ and its space is denoted as $\mathcal{Z} = \left\{ \mathcal{Z}_{1:t} \right\}$. Lastly, we omit the conditioning over future actions. The augmented MI  \eqref{eq:augmented_mi_full} then becomes
\begin{equation}
	\label{eq:mi_aug_def}
	\begin{aligned}
		\miaug{\X}{\Xnew}{\Z} \bydef \ent{\X} - \ent{\X' \mid Z} 
		 = \ent{\X} - \ent{\X, \Xnew \mid Z}.
	\end{aligned}
\end{equation}
The evaluation of any of the presented information-theoretic terms involves integration over the state-space, thus we might face the Curse of Dimensionality when the state is high-dimensional. When there are multiple tasks, then, arguably, evaluating the information-theoretic terms involves the heaviest calculations of the objective function.

\subsection{Non-parametric entropy estimation} \label{subsec:nonpara_ent_est}

For any of the presented information-theoretic terms, one possible calculation scheme is to go through calculation of entropy. Having  a non-parametric belief over a state $\X \in \mathbb{R}^D$, it is usually approximated by a weighted particle set $\{ (\X^{(i)}, w^{(i)}) \}_{i=1}^{N}$ with normalized weights, where $\X^{(i)}$ is the $i$-th particle, $w^{(i)}$ is the weight of the $i$-th particle, and $N$ is the number of particles. The Curse of Dimensionality in this case means that in order to have sufficient resolution to represent the belief, the number of particles $N$ needs to be exponential in the dimension $D$, i.e. $N \propto \alpha^D$ where $\alpha \geq 1$. The entropy in this case is approximated as well using this particle set, and thus will suffer from the Curse of Dimensionality as well, as will further be shown. 
Several entropy estimators exist in literature, each has its own advantages and disadvantages. The most well-known estimators can be found in \cite{Beirlant97jmss}. One of these estimators, for example, is the re-substitution estimator, for which the entropy is estimated as $\hat{\mathcal{H}} \left[ \X \right] \bydef \sum_{i=1}^{N} w^{(i)} \log \hat{b} \left[ \X^{(i)} \right]$,
where $\hat{b} \left[ \X \right]$ is an approximation of the belief obtained by a probability distribution estimator such as Kernel Density Estimator (KDE). The computational complexity of calculating entropy with the re-substitution estimator with KDE is $O \left( N^2 D \right)$.
Another estimator, presented in \cite{Stowell09spl}, performs $k$-d partitioning of the state-space, and thus achieves a complexity of $O \left( N \log N \right)$. There are many more estimators, such as estimators which are based on nearest-neighbor or $m_n$-spacings, however analyzing these is outside the scope of this work.

Although the complexity of the presented estimators might not seem to have an exponential relation to the state's dimension $D$, or any relation at all, we remind that $N$ should be exponential with $D$ in order to get an accurate enough estimation, thus the complexity of any estimator is exponential with the dimension $D$, even if not explicitly.


%% file: 04-Approach.tex

\graphicspath{ {./Figures/}}

Our work includes multiple contributions when approaching the problem of informative planning with high-dimensional non-parametric beliefs. These will be presented in different sections.
Our first contribution is an exact mathematical derivation which shows that the dimension of the belief-space can be reduced for the calculation of augmented MI by only exploiting the structure of the problem, which is true as well for the case of the original definition of MI. It thus provides a solution to the curse of dimensionality. We refer to this part of the approach as \invmi{}.
As our second contribution, we derive a method that allows to avoid the explicit reconstruction of the beliefs' surfaces usually required for the augmented MI calculation. We then use this to construct an estimator, which we refer to as \mismc{}.
In the last section, we discuss how \invmi{} and \mismc{} can be used together with existing solvers of the informative planning problem, more specifically tree-based solvers.

\subsection{Dimensionality reduction for MI calculation} \label{sec:dim_red}

We aim at reducing the complexity of calculating the augmented MI over a high-dimensional state $\X \in \mathbb{R}^D$, where $D \gg 1$, distributed with a non-parametric belief.
As the dimension increases, the number of samples required to get the same accuracy, and thus the complexity, exponentially increases.

More specifically, the \emph{key idea} of our approach is to reduce the dimensionality of the problem by exploiting its \emph{structure}. As in the works \cite{Kopitkov17ijrr}, \cite{Elimelech17icra}, it starts by dividing the high-dimensional \emph{prior} state into two subsets, such that $\X = \left\{ \Xinv, \Xnot \right\}$.
We remind that only subsets of the state participate in the probabilistic transition and observation models, $\mathbb{P}_T$ and $\mathbb{P}_Z$, respectively.
The same also applies for the sequential counterparts, $\FT$ and $\FZ$, defined earlier. Thus, $\Xinv \in \mathbb{R}^d$, which we will refer to as the \emph{involved} subset of the state, is defined as a union of all variables in the prior state $\X = \X_0 $ which participate in generating future state transitions and future observations
\begin{equation}
	\label{eq:inv_def}
	\Xinv = \left[ \bigcup_{i=1}^{t} \Big[ \X_{i-1}^{tr} \cup \X_{i}^{obs} \Big] \right] \cap \X,
\end{equation}
where the intersection with the prior state $\X$ is to emphasize that while the subsets $\X_{i-1}^{tr}$ and $\X_{i}^{obs}$ might also include new augmented variables $\Xnew$, which are inherently involved, we define $\Xinv$ as a subset of the prior alone.

Furthermore, this subset is usually very small compared to the dimension of the entire state, i.e. $d \ll D$, a fact which is of key importance in our approach.
Determining the involved subset is done heuristically, as is naturally done when transitioning and generating future measurements.
This idea is also illustrated with a SLAM example in Figure~\ref{fig:illustration_involved}. In this example, the involved variables are the prior pose and observed landmarks.

More generally, we define $\Xinvp$, a subset \emph{containing} the involved rather \emph{only} the involved, such that $\Xinv \subseteq \Xinvp \subseteq \X$. This subset has larger dimensions, yet it might still be much smaller in dimensions compared to the entire state. When referring to the involved variables, we also refer to this subset.

With the augmented MI definition in eq.~(\ref{eq:mi_aug_def}) we derive the following Lemma,

\begin{lem}
	\label{le:mi_miaug_relation}
	Let \(\X\) be some prior state, and \(\X' = \left\{ \X, \Xnew \right\}\) be a successor state, where \(\Xnew\) is the augmented subset of the successor state. Let the multivariate random variable \(\Z\) denote an observation sequence over this successor state. Then, the relation between MI and augmented MI is
	\begin{equation}
		\miaug{\X}{\Xnew}{\Z} = \mi{\X, \Xnew}{\Z} - \ent{\Xnew \mid \X}.
	\end{equation}
\end{lem}
where $\mi{\X, \Xnew}{\Z} \bydef \ent{\X, \Xnew} - \ent{\X, \Xnew \mid \Z}$ is the definition of the original MI over the joint state $\left\{ \X, \Xnew \right\}$.
The proof of this lemma, as well as the following theorems, lemmas and propositions, is given in Appendices \ref{sec:invmi} and \ref{sec:mismc}. 
In words, the difference between the two MI variants is the expected uncertainty obtained directly from transitioning the state.
Using this Lemma, we  derive the following Theorem,
\begin{theorem}
	\label{th:mi_aug_involved}
	Let \(\X\) be some prior state, and \(\X' = \left\{ \X, \Xnew \right\}\) be a successor state, where \(\Xnew\) is the augmented subset of the successor state. Let the multivariate random variable \(\Z\) be an observations sequence over a subset of the successor state, \( \left\{\Xinvp, \Xnew \right\} \subseteq \X'\), such that \( \prob{\Z \mid \X, \Xnew} \equiv \prob{\Z \mid \Xinvp, \Xnew} \). Then,
	\begin{equation}
		\miaug{\X}{\Xnew}{\Z} = \miaug{\Xinvp}{\Xnew}{\Z}.
	\end{equation}
\end{theorem}
%
In words, Theorem~\ref{th:mi_aug_involved} states that the expected information to be gained about the entire state between these time steps is exactly the expected information to be gained about the involved variables $\Xinvp$ alone. This, in turn, means that in order to get an \emph{exact} solution, it is sufficient to solve a much lower dimensional problem. We note that this is regardless of the correlations, which are already taken into account in the marginalization process.

The subset $\Xinv$ depends on the specific action and its observations, such that if we have $n$ candidate actions, each action might have different involved variables. We will denote the involved subset for the $j$-th action as $\Xinvi{j}$. Reasoning about the exact involved variables and marginalizing out the uninvolved variables for each action might be costly operations which will eventually make this whole approach worthless. When using the more general definition of the involved variables, $\Xinvp$, it allows choosing $\Xinvp = \Xinvi{1} \cup \Xinvi{2} \cup \dots \cup \Xinvi{n}$, which results in a one-time marginalization rather than marginalizing for each action separately. This concept is similar to the one suggested in \cite{Kopitkov17ijrr}. Moreover, using $\Xinvp$ might be helpful in cases we have an easy way to calculate marginalized beliefs of subsets which are not exactly the involved but contain them. Despite all this, for readability reasons we will continue to refer to the involved variables as $\Xinv$ rather than $\Xinvp$, yet we emphasize that the following is true for $\Xinvp$ as well.

\begin{figure}
	\centering
	\begin{subfigure}[b]{0.49\textwidth}
		\centering
		\scalebox{0.22}{
			\includegraphics{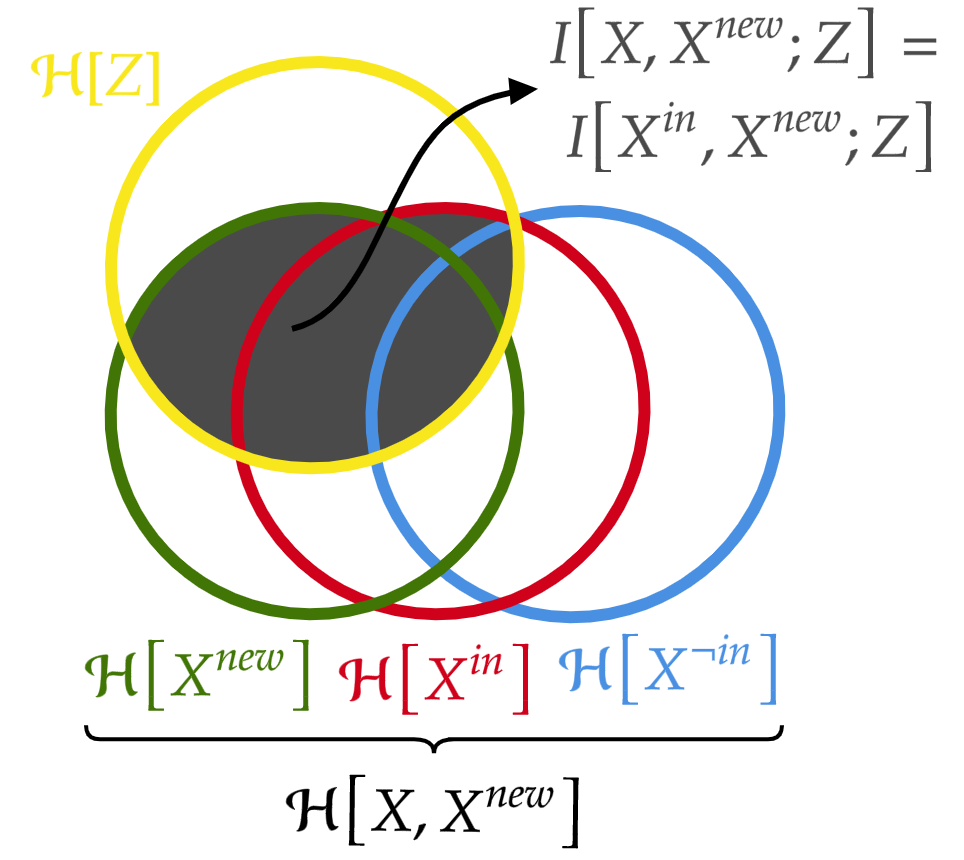}
		}
		\caption{}
		\label{fig:info_diag_1}
	\end{subfigure}
	\hfill
	\begin{subfigure}[b]{0.49\textwidth}
		\centering
		\scalebox{0.22}{
			\includegraphics{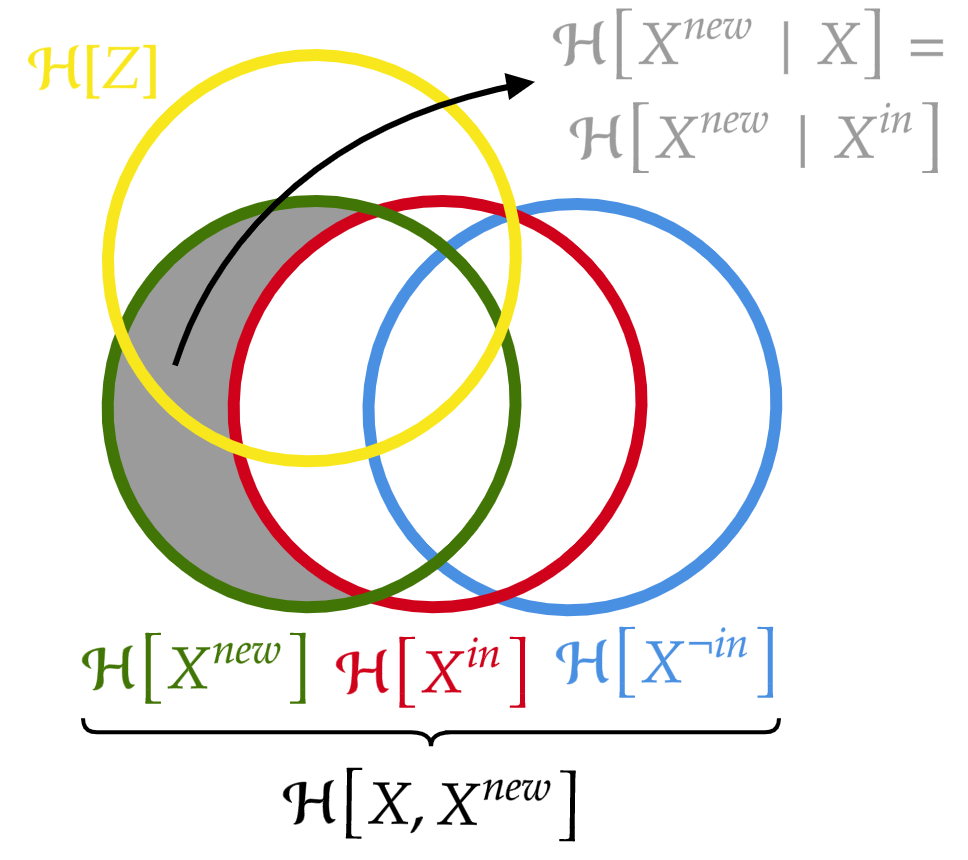}
		}
		\caption{}
		\label{fig:info_diag_2}
	\end{subfigure}
	\caption{Illustration of the augmented problem as an information diagram. The area of each circle represents the entropy value of a subset of variables. The mutual areas represent the MI values between these subsets. The key idea, obtained by Theorem~\ref{th:mi_aug_involved}, is that there is no mutual area between $\Xnot$ and both $\Xnew$, $\Z$ given $\Xinv$. The dark gray area in (a) is $\mi{\Xinv,\Xnew}{\Z}$, the light gray area in (b) is $\ent{\Xnew \mid \Xinv}$. Subtraction of the dark gray area by the light gray area yields the augmented MI. \vspace{-15pt}
	}
	\label{fig:augmented_info_diag}
\end{figure}

The result of Theorem~\ref{th:mi_aug_involved} is also illustrated as an information diagram in Figure~\ref{fig:augmented_info_diag}. Note that discarding (marginalizing out) the subset $\Xnot \bydef \X \setminus \Xinv$ (blue circle) does not affect the calculation since the shaded areas we are calculating remain the same.

While this conclusion might remind the one in \cite{Kopitkov17ijrr}, which considered only the Gaussian case, it can be viewed as a non-trivial extension to an arbitrary distribution. Specifically, Theorem~\ref{th:mi_aug_involved} states that \emph{in expectation} the augmented IG and its involved counterpart are exactly equal, no matter how the belief is distributed.

\begin{wrapfigure}{r}{7.5cm}
	\vspace{-20pt}
	\begin{minipage}{\textwidth}
		\begin{algorithm}[H]
			\caption{\invmi{}}
			\label{alg:invmi}
			\scriptsize
			\begin{algorithmic}[1]
				\Require $\bel{\X}$, $\FT$, $\FZ$, $a$
				\Ensure $\miaug{\X}{\Xnew}{\Z}$
				\State $\Xinv \gets$ \Call{DetermineInvolved}{$\bel{\X}, \FT, \FZ, a$}
				\State $\bel{\Xinv} \gets$ \Call{Marginalize}{$\bel{\X}, \Xinv$}
				\State $\miaug{\Xinv}{\Xnew}{\Z} \gets$ \Call{CalcMI}{$\bel{\Xinv}, \FT, \FZ, a$} 
				\State $\miaug{\X}{\Xnew}{\Z} \gets \miaug{\Xinv}{\Xnew}{\Z}$
			\end{algorithmic}
		\end{algorithm}
	\end{minipage}
	\vspace{-20pt}
\end{wrapfigure}

Using the result of Theorem \ref{th:mi_aug_involved}, we propose an approach which we will refer to as \invmi{}, and is summarized in Algorithm~\ref{alg:invmi}. Given the prior belief, the probabilistic models and an action sequence, this algorithm first determines the involved variables using some heuristic. Then, it calculates the marginalized prior belief over the involved variables. And, finally, it calculates the (augmented) MI value using the marginalized prior belief (while propagating future beliefs), using any calculation scheme. This is instead of naively using the entire prior belief.
We remind that either or both the complexity and the accuracy of any calculation scheme of the MI depend on the dimension of the entire state, $D$. The \emph{main contribution} of our approach is that it cancels this dependence. Instead, the dependence is over the dimension of a smaller subset of the state, $d$, which improves either or both the complexity and the accuracy.
In order to make the overall marginalization process more efficient, we can instead work with $\Xinvp$, and the algorithm is then slightly changed depending on how it is defined, yet the key idea remains the same.

Note also this approach is not limited to any specific calculation scheme of the MI. The calculation, for example, might go through its relation to entropy, by calculating the entropy terms or estimating them using entropy estimators, such as the common re-substitution estimator with KDE we have briefly introduced in Section \ref{sec:problem-formulation}.
We shall use it now to demonstrate the complexity reduction we get from using our approach. It is easy to first see that for the same number of samples, our approach reduces the complexity of estimating the entropy terms from $O \left( N^2 D \right)$ to $O \left( N^2 d \right)$. However, as we have already mentioned, the Curse of Dimensionality suggests that the number of samples required to get the same accuracy is exponential in the dimension. Thus, if we wish to preserve the accuracy, using our approach, only $n \propto \alpha^d$ samples are required, compared to $N \propto \alpha^D$ samples which were required without our approach. Since $d \ll D$, it means that $n \ll N$. This reduces the complexity even further to $O \left( n^2 d \right)$. To generalize and simplify our conclusion, we consider the complexity of any estimator is at least linear in the number of samples, thus the complexity is reduced, at least, from $O \left( \alpha^{D} \right)$ to $O \left( \alpha^{d} \right)$, where we remind that $d \ll D$ and $\alpha \geq 1$.

\subsection{Avoiding the reconstruction of future beliefs' surfaces} \label{subsec:no_kde}

The estimation scheme of the involved MI might require reconstructing the surfaces of future beliefs, which adds to the estimation error or perhaps entails another level of complexity in the form of new hyperparameters.
In this section, we present a theoretical derivation that allows to avoid this reconstruction step for the augmented MI calculation. We then use this derivation to construct an estimator.
This estimator, however, as will shortly be explained, can only be used in cases in which the probabilistic models are explicitly given. 
\begin{theorem}
	\label{th:mi_augmented_superposition}
	Let \(\X\) be some prior state, and \(\X' = \left\{ \X, \Xnew \right\}\) be a successor state, where \(\Xnew\) is the augmented subset of the successor state. Let \(\Z\) be an observation sequence over a subset of the successor state, \( \left\{\Xinv, \Xnew \right\} \subseteq \X'\), such that \( \prob{\Z \mid \X, \Xnew} \equiv \prob{\Z \mid \Xinv, \Xnew} \). Then, the augmented MI can be factorized as
	\begin{equation}
		\label{equ:mi_augmented_superposition}
		\miaug{\X}{\Xnew}{\Z} = -\ent{\Xnew \mid \Xinv} - \ent{\Z \mid \Xinv, \Xnew} + \ent{\Z}.
	\end{equation}
\end{theorem}
This result suggests that the augmented MI can be expressed as a superposition of the expected entropies of the sequential transition model, observation model and normalizer. More explicitly, the augmented MI can be written as
\begin{equation}
	\label{eq:mi_augmented_integrals}
	\begin{aligned}
		\miaug{\X}{\Xnew}{\Z} &=
		\int_{\XinvSp} \bel{\Xinv} \left[ \int_{\XnewSp} \FT  \log \FT d\Xnew \right] d\Xinv \\
		&+ \int_{\XinvSp} \bel{\Xinv} \Bigg[ \int_{\XnewSp} \FT 
		\left[ \int_{\ZSp} \FZ \log \FZ d\Z \right] d\Xnew \Bigg] d\Xinv \\
		&- \int_{\XinvSp} \bel{\Xinv} \bigg[ \int_{\XnewSp} \FT 
		\left[ \int_{\ZSp} \FZ \log \eta^{-1} d\Z \right] d\Xnew \bigg] d\Xinv,
	\end{aligned}
\end{equation}
where the normalizer can be calculated with
\begin{equation}
	\label{eq:normalizer}
	\eta^{-1} =
	\int_{\XinvSp} \bel{\Xinv} \left[ \int_{\XnewSp} \FT \FZ d\Xnew \right] d\Xinv.
\end{equation}
Eqs.~\eqref{eq:mi_augmented_integrals} and \eqref{eq:normalizer} suggest that the objective function can be calculated without the need to reconstruct future beliefs' surfaces, which is a \emph{key result}. Also, the integration is over the involved and new variables, already exploiting the dimensionality reduction.

We then use sampling to estimate eqs.~\eqref{eq:mi_augmented_integrals} and \eqref{eq:normalizer}, as summarized  in Algorithm~\ref{alg:invmi-smc}. The full details can be found in Appendix \ref{sec:Developing-estimator}. 
Since it uses particles from the prior and propagates them as in Sequential Monte Carlo (SMC) methods, we will refer to it as \mismc{}.
Note that this algorithm can be an \emph{anytime} algorithm, since the calculation can be updated incrementally when adding more particles.


Also note that although this estimator is formulated by already exploiting the dimensionality reduction, it is not vital. The estimator can get as an input the full prior belief $\bel{\X}$ instead of the involved prior belief $\bel{\Xinv}$. The particles would then be high-dimensional, yet since these are used only for the evaluation of the probabilistic models, marginalization would automatically be done in the context of these particles. This attribute makes this estimator closely related to the \invmi{} approach, without explicitly using it beforehand. However, to avoid the Curse of Dimensionality, it is preferred to maintain and sample from a lower-dimensional belief to begin with, which is exactly the result of using \invmi{}.


\begin{wrapfigure}{r}{6.8cm}	
	\vspace{-35pt}
	\begin{minipage}{0.95\textwidth}
		\begin{algorithm}[H]
			\caption{\mismc}
			\label{alg:invmi-smc}
			\scriptsize
			\begin{algorithmic}[1]
				\Require $\bel{\Xinv}$, $\FT$, $\FZ$, $a$
				\Ensure $\miaug{\Xinv}{\Xnew}{\Z}$
				\State $sum_1 \gets 0$
				\State $sum_2 \gets 0$
				\State $sum_3 \gets 0$
				\For{$i=1$ \textbf{to} $n_1$}
				\State $\left(\xinv{i},w^{(i)} \right) \sim \bel{\Xinv}$
				\For{$j=1$ \textbf{to} $n_2$}
				\State $\xnew{i,j} \sim \FT(\Xnew \mid \xinv{i}, a)$
				\State $value_1 \gets w^{(i)} \frac{1}{n_2} \log \FT^{(i,j)}$
				\State $sum_1 \gets sum_1 + value_1$
				\For{$k=1$ \textbf{to} $n_3$}
				\State $\zz{i,j,k} \sim \FZ(\Z \mid \xinv{i}, \xnew{i,j})$
				\State $value_2 \gets w^{(i)} \frac{1}{n_2} \frac{1}{n_3} \log \FZ^{(i,j,k)}$
				\State $sum_2 \gets sum_2 + value_2$
				\State $\eta^{-1} \gets 0$
				\For{$l=1$ \textbf{to} $n_4$}
				\State $\left(\xinv{l},w^{(l)} \right) \sim \bel{\Xinv}$
				\For{$m=1$ \textbf{to} $n_5$}
				\State $\xnew{l,m} \sim \FT(\Xnew \mid \xinv{l}, a)$
				\State $value \gets w^{(l)} \frac{1}{n_5} \FZ^{(l,m,k)}$
				\State $\eta^{-1} \gets \eta^{-1} + value$
				\EndFor
				\EndFor
				\State $value_3 \gets w^{(i)} \frac{1}{n_2} \frac{1}{n_3} \log \eta^{-1}$
				\State $sum_3 \gets sum_3 + value_3$
				\EndFor
				\EndFor
				\EndFor
				\State $\miaug{\Xinv}{\Xnew}{\Z} \gets sum_1 + sum_2 - sum_3$
			\end{algorithmic}
		\end{algorithm}
	\end{minipage}
	\vspace{-20pt}
\end{wrapfigure}
Considering that we have a total number of $m$ observations instances and a total number of $n$ particles the complexity becomes $O \left( m n d \right)$. In comparison, the complexity of using a re-substitution estimator with KDE is $O \left( m n^2 d \right)$ when using \invmi{}, which makes our estimator favorable in terms of complexity. The full analysis can be found in Appendix \ref{sec:Complexity}. 
We also again emphasize that our estimator avoids the intermediate step of belief surface reconstruction, and hence we conjecture it is expected to be more accurate.

We remind that many more other estimators exist in the literature, and we show in Appendix \ref{sec:Complexity} that \mismc{} is comparable to two of them. Further comparison to additional estimators is left for future research.



\subsection{Applicability to Belief Trees}

Up to this point, we have shown efficient ways of calculating augmented MI, which is an expectation over the reward, augmented IG. In general, the objective function can then be evaluated through these expected rewards due to commutativity. However, a common solving method, which is to construct a search over a belief-tree, goes through a direct calculation of the reward. The planning literature contains lots of tree-based solvers (e.g. \cite{Sunberg18icaps}, \cite{Fischer20icml}) with which \invmi{} and \mismc{} should be able to cope. Fortunately, we have found that we can define two new rewards, both of which already using \invmi{}, which makes this possible. Meaning, using these new rewards generate the same optimization problem.
For full details and proofs, the reader is refereed to Appendix \ref{sec:AppBeliefTrees}. 

%% file: 05-Results.tex
\graphicspath{ {./Figures/}}


Our approach was tested on an instance of active Simultaneous Localization and Mapping (SLAM) problem, which is a classical choice for high-dimensional problems, since the state contains past trajectory and the map. As in Figure \ref{fig:slam_involved}, an autonomous drone is flying around, observing landmarks which construct a 2D map. At each time step, it needs to decide where to move next in order to reduce its state uncertainty, i.e. the drone's trajectory and the map. In order to make a decision, the drone estimates the augmented MI of the different possible actions at that time. For states which are distributed with Gaussians, the augmented MI can be evaluated using an analytical solution, which makes it a perfect choice as a first validation of our approach. Although Gaussian, we emphasize that the different algorithms get samples as an input, thus work as if it is a purely non-parametric scenario. Using this simulation, we have tested (i) the impact of the dimensionality on choosing an action; and (ii) the impact of increasing the dimensionality on accuracy and timing.

\subsection{Impact of the dimensionality on choosing an action}

At time $t$, the drone needs to choose between four different actions, each involving an observation of a different landmark. In this section, we compare between the analytical solution and the estimation results obtained by three different methods, which we will refer to as \nkde{}, \invmikde{} and \invmismc{}. \nkde{} is the naive approach, which uses a re-substitution estimator with KDE over the entire state; \invmikde{} first uses the \invmi{} approach and only then the re-substitution estimator with KDE, i.e. the estimation is only over the involved subset of the state; \invmismc{} is our suggested estimation scheme, \mismc{}, where we just emphasize that \invmi{} is inherent in it.
We note that for these specific tests, the methods which included KDE were implemented as if the inference engine is perfect, i.e. the posterior samples were generated from the true posterior, whereas \mismc{}, as shown in Algorithm~\ref{alg:invmi-smc}, uses samples from the prior belief and propagates them in a SMC manner. This gives a slight advantage to the methods with KDE over \invmismc{}.
The simulations were done in Python, where we used Scipy's KDE for the two KDE methods.

\begin{figure}
	\centering
	\caption{A comparison between the mean MI values of the different actions and calculation methods. The standard deviation of the calculation methods are shown as error bars. While the mean values, even if not close to the analytical values, maintain action consistency, the real problem is the standard deviation. For a specific trial, the actions ordering might be changed due to the overlaps between MI's possible values. The naive approach suffers the most from this problem.
}
	\scalebox{0.35}{
		\includegraphics{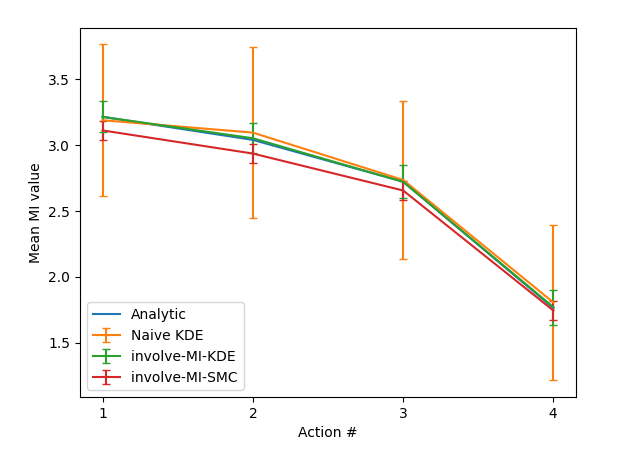}}
	\label{fig:actions_compare}
\end{figure}

The prior state has $\sim 150$ dimensions, whereas each action involves a subset of the state with only $4$ dimensions. Each method was provided with $300$ particles (for \invmismc{}, $n_1 = n_4 = 300$, $n_2 = n_3 = n_5 = 1$), and the MI was calculated $100$ times to evaluate its standard deviation. The mean values and standard deviations of each method and for each action are shown in Figure \ref{fig:actions_compare}.
As expected, \nkde{} has big mean errors and big variances. These errors are big enough to make the drone choose an action which isn't optimal. \invmikde{} and \invmismc{}, on the other hand, both present pretty similar and better results compared with \nkde{}, with the former being slightly better in terms of the mean value, probably due to the prior advantage we have provided it with. We note that also for the case of the \invmi{} estimators, the drone might choose an action which isn't optimal due to the similarity between the MI values of this specific test, however an increase in the number of samples will solve this (a smaller increase compared with the one needed for the naive approach).

\subsection{Impact of increasing the dimensionality on accuracy and timing}

\begin{figure}
	\centering
	\begin{subfigure}[b]{0.49\textwidth}
		\centering
		\scalebox{0.35}{
			\includegraphics{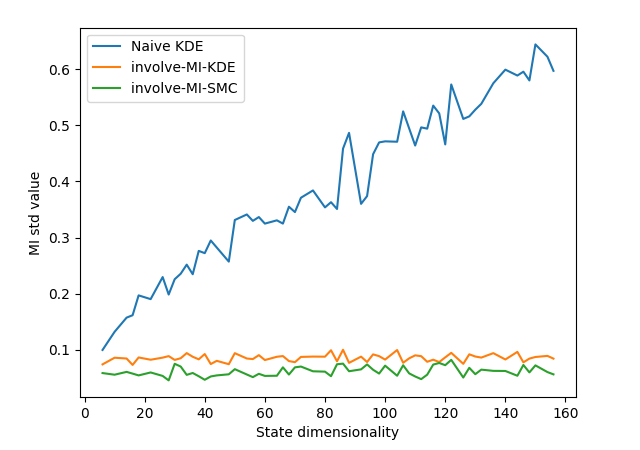}
		}
		\caption{}
		\label{fig:std_with_dimension}
	\end{subfigure}
	\hfill
	\begin{subfigure}[b]{0.49\textwidth}
		\centering
		\scalebox{0.35}{
			\includegraphics{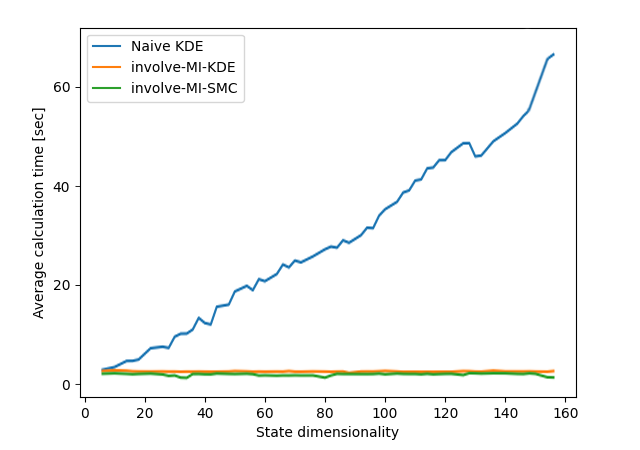}
		}
		\caption{}
		\label{fig:timing_with_dimension}
	\end{subfigure}
	
	\caption{Impact of dimensionality on MSE and calculation time for the three methods. (a) shows the standard deviation, while (b) shows the average calculation time. {\color{blue} \nkde{}}'s error and calculation time both linearly increase with the state's dimension. Using \invmi{} keeps both roughly constant, independent of the state's dimension.
	}
	\label{fig:dimenson_impact}
	\vspace{-5pt}
\end{figure}

We have also tested the impact of increasing state dimensionality on the standard deviation of each estimator, where the action now stays the same.
The original motivation is to show that as the dimension grows, an exponential number of samples would be required in order to get the same accuracy for the naive approach. However, it is not feasible with the dimensions we chose for this simulation. Instead, we approach it in a different manner, and show that as the dimension grows, for a constant number of samples, the accuracy is harmed.
Each estimator, again, was provided with $300$ particles and the MI was calculated $100$ times to evaluate its standard deviation and the average calculation time. The results can be seen in Figure \ref{fig:dimenson_impact}. Both the standard deviation and calculation time of the MI using \nkde{} increase with the state's dimension, an increase which seems linear. On the other hand, for \invmikde{} and \invmismc{} both the standard deviation and calculation time are roughly constant. This is thanks to the fact that the involved subset is of the same dimension during this simulation (only one landmark is observed for each action). This demonstrates our main contribution, for which \invmi{} is better both in terms of accuracy and time complexity, for the same number of samples.
The standard deviation and timing of both \invmikde{} and \invmismc{} are comparable, with a slight advantage to \invmismc{}. We remind that \invmikde{} has a complexity of $O (m n^2 d)$ whereas \invmismc{} has a complexity of $O (m n d)$, which suggests that a better performance should have been obtained for \invmismc{}. However, it is very likely that the Scipy's KDE implementation is optimized, whereas the implementation of \invmismc{} is currently very simple and straightforward. We thus conjecture that the timing could be further improved. We also conjecture that providing the KDE methods with a practical inference engine, rather than the current perfect inference it was provided with, and using more complex distributions, will increase the gap in terms of accuracy.

%

%% file: 06-Conclusions.tex
%
%

To conclude, we have identified a void in the informative planning subject. For the case of high-dimensional non-parametric beliefs, the complexity becomes too high for solving (not to mention online solutions), thus current state-of-the-art approaches either avoid the high-dimensionality, assume very specific settings or apply rough approximations which impact the accuracy.

In this work, we have filled this void, mainly by reducing the dimensionality of the state for the expected reward's (augmented MI) calculation, while relaxing many of the former assumptions and approximations.
Next, we have introduced an estimator, \mismc{}, which avoids the reconstruction of future belief's surfaces in case that we have access to the probabilistic models of the problem, which we conjecture might reduce the estimation error and timing, compared to KDE-based approaches, for example. This, in turn, also makes it an anytime algorithm.
Lastly, for the completeness of this work, we have related back to the planning optimization problem and shown how both \invmi{} and \mismc{} can be applied in it.
We emphasize again that this work contributes specifically to the objective function calculation, and thus can be used as a black box together with many other state-of-the-art algorithms which contribute to other building blocks of the planning task.


In future work, we would like to extend our approach to the more complex focused case, for which we are only interested in quantifying the uncertainty over a subset of the entire state, as it was done in \cite {Kopitkov17ijrr} for the Gaussian case. Another key aspect that we would like to further investigate is the impact of the non-parametric inference engine in this context. Given the latest novelties in this field, such as the work of \cite{Huang21icra}, we believe it would be very relevant for efficient and accurate non-parametric marginalization.



%% file: 07-Appendix.tex
\section{Information-theoretic rewards}\label{sec:Info_theo}

In this section we provide the definitions of the original Information Gain (IG) and Mutual Information (MI). IG quantifies the amount of information gained for a certain variable $\X$ (the state) by knowing the value of another variable $\Z$ (an observation). It is defined as the difference between the entropy prior to this additional knowledge and the entropy afterwards
\begin{equation}
\ig{X}{Z=z} \bydef \ent{\X} - \ent{\X \mid \Z=z}.
\end{equation}
MI is IG in expectation, and it can also be defined as the difference between the entropy of the state and the expected entropy of the state given an observation
\begin{equation}
\begin{aligned}
\mi{\X}{\Z}
\bydef \underset{\mathcal{Z}}{\mathbb{E}} \Big[ \ig{X}{Z=z} \Big]
= \ent{\X} -\ent{\X \mid \Z}.
\end{aligned}
\label{eq:MI}
\end{equation}
where $\ent{\X \mid \Z} = \expec{\ZSp} \big[ \ent{\X \mid \Z=z} \big]$. For the case where the state changes between time steps, as discussed in the paper, the original definitions of IG and MI are insufficient, since these do not account for the additional uncertainty obtained by changes in the state.


\section{\invmi{}}\label{sec:invmi}

\subsection{Proof of Lemma 1}
We remind the augmented MI is
\begin{equation}
	\miaug{\X}{\Xnew}{\Z} \bydef \ent{\X} - \ent{\X, \Xnew \mid Z}.
\end{equation}
Using the following known identities
\begin{align}
	\ent{\X} &\bydef \ent{\X, \Xnew} - \ent{\Xnew \mid \X} \\
	\mi{\X, \Xnew}{\Z} &\bydef \ent{\X, \Xnew} - \ent{\X, \Xnew \mid Z},
\end{align}
we get the relation between MI and the augmented MI
\begin{equation}
	\label{eq:mi_miaug_relation}
	\miaug{\X}{\Xnew}{\Z} = \mi{\X, \Xnew}{\Z} - \ent{\Xnew \mid \X}.
\end{equation}


\subsection{Proof of Theorem 1}
Using the chain rule for MI, where the state is arbitrarily partitioned as $\X' = \{ \X^A, \X^B \}$ yields
\begin{equation}
	\label{eq:chain_rule}
	\begin{aligned}
		\mi{\X'}{\Z}
		= \mi{\X^A, \X^B}{\Z}
		= \mi{\X^A}{\Z} + \micond{\X^B}{\Z}{\X^A}.
	\end{aligned}
\end{equation}
By definition, the conditional MI term is
\begin{equation}
	\label{eq:micond_def}
	\begin{aligned}
		\micond{\X^B}{\Z}{\X^A} \bydef
		&\int_{\Z} \int_{\X^A} \int_{\X^B} \prob{\Z, \X^A, \X^B} \cdot \\
		&\cdot \log \left[ \frac{\prob{\X^B, \Z \mid \X^A}}{\prob{\X^B \mid \X^A} \prob{\Z \mid \X^A}} \right] d\X^B d\X^A d\Z.
	\end{aligned}
\end{equation}
Using the chain rule over the numerator inside the $\log$ term, we get
\begin{equation}
	\label{eq:joint_xnot_z}
	\prob{\X^B, \Z \mid \X^A} = \prob{\Z \mid \X^A, \X^B} \prob{\X^B \mid \X^A}.
\end{equation}
Defining $\X^A \bydef \left\{ \Xinvp, \Xnew \right\}$, meaning it contains all the variables involved in generating the observations $\Z$ (and all new states $\Xnew$), we can state that $\prob{\Z \mid \X^A, \X^B} = \prob{\Z \mid \X^A}$, so eq.~(\ref{eq:joint_xnot_z}) becomes
\begin{equation}
	\prob{\X^B, \Z \mid \X^A} = \prob{\Z \mid \X^A} \prob{\X^B \mid \X^A}.
\end{equation}
Plugging this term back into eq.~(\ref{eq:micond_def}) yields
\begin{equation}
	\micond{\X^B}{\Z}{\X^A} \bydef
	\int_{\Z} \int_{\X^A} \int_{\X^B} \prob{\Z, \X^A, \X^B}
	\log \left( 1 \right) d\X^B d\X^A d\Z = 0.
\end{equation}
Using the above result, eq.~(\ref{eq:chain_rule}) then transforms into
\begin{equation}
	\label{eq:successor_mi_involved}
	\mi{\X'}{\Z} = \mi{\Xinvp, \Xnew}{\Z}.
\end{equation}
With our definition of $\X^A$, the prior state can be written as $\X = \left\{ \Xinvp, \X^B \right\}$.
Looking then at the conditional entropy term in the result of Lemma 1, we can rewrite it as $\ent{\Xnew \mid \X} = \ent{\Xnew \mid \Xinvp, \X^B}$. By our definition of $\Xinvp$, $\Xnew$ is conditionally independent of $\X^B$ given $\Xinvp$, i.e. $\prob{\Xnew \mid \Xinvp, \X^B} = \prob{\Xnew \mid  \Xinvp}$. Thus, one of the conditional entropy properties states that
\begin{equation}
	\label{eq:cond_indep}
	\ent{\Xnew \mid \X} = \ent{\Xnew \mid \Xinvp}
\end{equation}
Plugging (\ref{eq:successor_mi_involved}) and (\ref{eq:cond_indep}) back into the result of Lemma 1 (eq.~(\ref{eq:mi_miaug_relation})) we get that
\begin{equation}
	\begin{aligned}
			\miaug{\X}{\Xnew}{\Z} =
			\mi{\Xinvp, \Xnew}{\Z} - \ent{\Xnew \mid \Xinvp}.
		\end{aligned}
	\label{eq:mi_aug_involved_pre}
\end{equation}
We then observe that by using the result from eq.~({\ref{eq:mi_miaug_relation}}), the right hand side in eq.~({\ref{eq:mi_aug_involved_pre}}) is equal to $\miaug{\Xinvp}{\Xnew}{\Z}$, and so we finally conclude that
\begin{equation}
	\label{eq:mi_aug_involved}
	\miaug{\X}{\Xnew}{\Z} = \miaug{\Xinvp}{\Xnew}{\Z}.
\end{equation}


\section{\mismc{}}\label{sec:mismc}

\subsection{Proof of Theorem 2}
We begin by using the definition of the augmented MI over the involved subset, which is
\begin{equation}
	\label{eq:mi_aug_inv_def}
	\miaug{\Xinv}{\Xnew}{\Z} \bydef \ent{\Xinv} - \ent{\Xinv, \Xnew \mid \Z},
\end{equation}
where we remind that we use $\Xinv$ instead of $\Xinvp$ for the readability of the paper, yet the analysis is true for the more general subset $\Xinvp$. Using the chain rule for conditional entropy over the second term on the right hand side of eq.~(\ref{eq:mi_aug_inv_def}) yields
\begin{equation}
\ent{\Xinv, \Xnew \mid \Z} = \ent{\Xinv, \Xnew, \Z} - \ent{\Z}.
\end{equation}
Using the same principle twice again eventually yields
\begin{equation}
\begin{aligned}
	\ent{\Xinv, \Xnew \mid \Z} =
	\ent{\Xinv} + \ent{\Xnew \mid \Xinv} 
	+ \ent{\Z \mid \Xinv, \Xnew} - \ent{\Z}.
\end{aligned}
\end{equation}
Plugging back into eq.~(\ref{eq:mi_aug_inv_def}), we observe that the term $\ent{\Xinv}$ is canceled out. Then, by using the result of Theorem 1, given in eq.~(\ref{eq:mi_aug_involved}), the augmented MI term over the high-dimensional state finally becomes
\begin{equation}
\label{eq:mi_augmented_superposition}
\begin{aligned}
	\miaug{\X}{\Xnew}{\Z} =
	- \ent{\Xnew \mid \Xinv}
	- \ent{\Z \mid \Xinv, \Xnew}
	+ \ent{\Z}.
\end{aligned}
\end{equation}


\subsection{Developing the estimator}\label{sec:Developing-estimator}

We remind that the augmented MI can be written as
\begin{equation}
	\label{eq:mi_augmented_integrals}
	\begin{aligned}
		\miaug{\X}{\Xnew}{\Z} &=
		\int_{\XinvSp} \bel{\Xinv} \left[ \int_{\XnewSp} \FT  \log \FT d\Xnew \right] d\Xinv \\
		&+ \int_{\XinvSp} \bel{\Xinv} \Bigg[ \int_{\XnewSp} \FT 
		\left[ \int_{\ZSp} \FZ \log \FZ d\Z \right] d\Xnew \Bigg] d\Xinv \\
		&- \int_{\XinvSp} \bel{\Xinv} \bigg[ \int_{\XnewSp} \FT 
		\left[ \int_{\ZSp} \FZ \log \eta^{-1} d\Z \right] d\Xnew \bigg] d\Xinv,
	\end{aligned}
\end{equation}
where the normalizer can be calculated with
\begin{equation}
	\label{eq:normalizer}
	\eta^{-1} =
	\int_{\XinvSp} \bel{\Xinv} \left[ \int_{\XnewSp} \FT \FZ d\Xnew \right] d\Xinv.
\end{equation}
We then approach to sampling, i.e.
\begin{equation}
\begin{gathered}
	\left( \xinv{i}, w^{(i)} \right) \sim \bel{\Xinv} \\
	\xnew{i,j} \sim \FT \left( \Xnew \mid \xinv{i} \right) \\
	\zz{i,j,k} \sim \FZ \left( \Z \mid \xinv{i}, \xnew{i,j} \right),
\end{gathered}
\end{equation}
which allows the augmented MI to be approximated as
\begin{equation}
\begin{gathered}
	\miaug{\X}{\Xnew}{\Z} \approx
	\sum_{i=1}^{n_1} w^{(i)} \left[ \frac{1}{n_2} \sum_{j=1}^{n_2} \log \FT^{(i,j)} \right] \\
	+ \sum_{i=1}^{n_1} w^{(i)} \left[ \frac{1}{n_2} \sum_{j=1}^{n_2} \left[ \frac{1}{n_3} \sum_{k=1}^{n_3} \log \FZ^{(i,j,k)} \right] \right] \\
	- \sum_{i=1}^{n_1} w^{(i)} \left[  \frac{1}{n_2} \sum_{j=1}^{n_2} \left[ \frac{1}{n_3} \sum_{k=1}^{n_3} \log {\eta^{-1}}^{(i,j,k)} \right] \right],
\end{gathered}
\end{equation}
where
\begin{equation}
\begin{gathered}
	\FT^{(i,j)} = \FT \left( \xinv{i}, \xnew{i,j} \right) \\
	\FZ^{(i,j,k)} = \FZ \left( \xinv{i}, \xnew{i,j}, \zz{i,j,k} \right) \\
	{\eta^{-1}}^{(i,j,k)} = \eta^{-1} \left( \xinv{i}, \xnew{i,j}, \zz{i,j,k} \right),
\end{gathered}
\end{equation}
and the normalizer, for each sampled instance, is then also approximated as
\begin{equation}
{\eta^{-1}}^{(i,j,k)} \approx
\sum_{l=1}^{n_4} w^{(l)} \left[ \frac{1}{n_5} \sum_{m=1}^{n_5} \FZ^{(l,m,k)} \right].
\end{equation}
\emph{Remark}: As in a particle filter, $\FZ^{(l,m,k)}$ can be considered an update for the particle's weight. Thus, the approximation of ${\eta^{-1}}^{(i,j,k)}$ can be viewed as an average of the updated weights.


\subsection{Complexity}\label{sec:Complexity}

\subsection*{Complexity analysis}

In terms of complexity, the most expensive step of this approach is the estimation of $\ent{\Z}$, thus its complexity is the complexity of the entire estimator. Estimating each ${\eta^{-1}}^{(i,j,k)}$ has a complexity of $O \left( n_4 n_5 d \right)$. In turn, the complexity of estimating $\ent{\Z}$ is of $O \left( n_1 n_2 n_3 n_4 n_5 d \right)$. Considering that we have a total number of $m$ observations instances, i.e. $n_1 n_2 n_3 = m$, and also that the total number of particles is $n$, i.e. $n_4 n_5 = n$, the complexity becomes $O \left( m n d \right)$.

\subsection*{In comparison to two more estimators}

Many other entropy estimators exist in the literature, such as the nearest neighbor estimator, which can be found in \cite{Beirlant97jmss}, and the $k$-d partitioning estimator, presented in \cite{Stowell09spl}. When estimating the MI value with these estimators, the complexity of both can get to $O \left(m n \log n \right)$, which is comparable to the complexity of \mismc{} when reminding that $n$ should be exponential in the dimension $d$.


\section{Applicability to belief trees}\label{sec:AppBeliefTrees}
In this section we wish to relate the approaches in the paper to the informative planning optimization problem.
We remind that although the following analysis considers an open-loop formulation, for which we seek for an optimal action sequence, $a_{0:T-1}$, it also applies for a close-loop formulation, for which we seek for a policy, $\pi_{0:T-1}$.
The solution to the $\rho$-POMDP is obtained by maximization of the objective function, denoted shortly as $J_0 \bydef J \left( \bel{\X_0}, a_{0:T-1} \right)$
\begin{equation}
	\label{eq:optimal_objective}
	J_0^{\star} = \underset{a_{0:T-1}}{\max} \left\{ \expec{\ZSp_{1:T}} \left[ \sum_{t=0}^{T-1} \rho_t +\rho_T \right] \right\}.
\end{equation}
Formulating it recursively yields the Bellman optimality equation
\begin{equation}
	\label{eq:bellman_optimality}
	J_t^{\star} = \underset{a_t}{\max} \left\{ \rho_t + \expec{\ZSp_{t+1}} \left[ J_{t+1}^{\star} \right] \right\}.
\end{equation}
where $J_t \bydef J \left( \bel{\X_t}, a_{t:T-1} \right)$.

A common solver to this optimization problem is to construct a search over a tree. More specifically, for $\rho$-POMDP, which is the case of belief-dependent rewards, a belief tree is used. In a belief tree, the beliefs $\bel{\X_t}$ are propagated using instances of future actions and observations, then the rewards $\rho_t$ are calculated, and
the action sequence providing the maximum value for the objective function is eventually chosen.
Since, in general, the action and observation spaces can be large, in order to be able to solve this optimization problem in reasonable time, it is approximated with a belief tree which propagates only a few sampled instances of future actions and observations. Dealing with continuous such spaces, a belief tree is an approximation of the problem to begin with.
The planning literature contains lots of tree-based solvers. However, since our analysis so far was done considering an expected reward, augmented MI, it is not trivial to prove that our approach, \invmi{}, and our estimator, \mismc{}, are able to cope with such solvers. This is the purpose of this section.

We denote the augmented IG, the augmented MI and their involved counterparts shortly as
\begin{align*}
	\igseq &\bydef \igaug{\X_0}{\x_{1:t}}{\Z_{1:t} = \z_{1:t} \mid a_{0:t-1}} \\
	\miseq &\bydef \miaug{\X_0}{\x_{1:t}}{\Z_{1:t} \mid a_{0:t-1}} \\
	\igseqinv &\bydef \igaug{\X_0^{in}}{\x_{1:t}}{\Z_{1:t} = \z_{1:t} \mid a_{0:t-1}} \\
	\miseqinv &\bydef \miaug{\X_0^{in}}{\x_{1:t}}{\Z_{1:t} \mid a_{0:t-1}}.
\end{align*}
We will also from now omit the term "augmented" while still referring to the more general case of augmentation. For readability, our analysis is done for IG as the only term of the reward, meaning $\rho_t = IG_0^t$, $\forall$ $t \in \left[ 1,T \right]$. Yet, the conclusions will also apply when there are additional terms for the reward, state-based terms for example.
Eq.~(\ref{eq:optimal_objective}) then becomes
\begin{equation}
	J_0^{\star} = \underset{a_{0:T-1}}{\max} \left\{ \sum_{t=0}^{T} \expec{\ZSp_{1:T}} \left[ \igseq \right] \right\}
	= \underset{a_{0:T-1}}{\max} \left\{ \sum_{t=0}^{T} \miseq \right\}.
\end{equation}
Using Theorem 1 over this equation yields
\begin{equation}
	\label{eq:objective_mi_involved}
	J_0^{\star} = \underset{a_{0:T-1}}{\max} \left\{ \sum_{t=0}^{T} \miseqinv \right\}.
\end{equation}

\begin{theorem}
	\label{th:involved-reward}
	Let us define a new reward, \( \rho_t^{in} = \igseqinv \). Solving the \( \rho \)-POMDP optimization problem with this reward is equivalent to solving it with the original reward, \( \rho_t = \igseq \),
	such that
	\begin{equation}
		J_t^{\star} = \underset{a_t}{\max} \left\{ \rho_t^{in} + \expec{\ZSp_{t+1}} \left[ J_{t+1}^{\star} \right] \right\}.
	\end{equation}
\end{theorem}


	\subsection*{Proof}
	We remind eq.~(\ref{eq:objective_mi_involved}) is
	\begin{equation}
		\label{eq:objective_mi_involved_remind}
		J_0^{\star} = \underset{a_{0:T-1}}{\max} \left\{ \sum_{t=0}^{T} \miseqinv \right\}.
	\end{equation}
	The involved MI is by definition an expectation over the involved IG
	\begin{equation}
		\miseqinv \bydef \expec{\ZSp_{1:T}} \left[ {IG_0^t}^{in} \right].
	\end{equation}
	Plugging this back into eq.~(\ref{eq:objective_mi_involved_remind}) yields
	\begin{equation}
		J_0^{\star} = \underset{a_{0:T-1}}{\max} \left\{ \sum_{t=0}^{T} \left[ \expec{\ZSp_{1:t}} \left[ \igseqinv \right] \right] \right\}.
	\end{equation}
	Due to commutativity, we can again switch between the order of expectation and summation, which yields
	\begin{equation}
		J_0^{\star} = \underset{a_{0:T-1}}{\max} \left\{ \expec{\ZSp_{1:T}} \left[ \sum_{t=0}^{T} \left[ \igseqinv \right] \right] \right\}.
	\end{equation}
	We then separate the first action $a_0$ from the rest of the actions $a_{1:T-1}$. We also observe that $\Z_1$ is not a function of $a_{1:T-1}$ and that $IG{_0^0}^{in}$ is not a function of both $a_{1:T-1}$ and $\Z_1$. This yields
	\begin{equation}
		J_0^{\star} = \underset{a_0}{\max} \left\{ IG{_0^0}^{in} + \expec{\ZSp_1} \left[ \underset{a_{1:T-1}}{\max} \left\{ \expec{\ZSp_{2:T}} \left[ \sum_{t=1}^T \igseqinv \right] \right\} \right] \right\}.
	\end{equation}
	We then observe that the term inside the expectation over $\Z_1$ is equal to $J_1^{\star}$, which yields the following recursive form
	\begin{equation}
		J_0^{\star} = \underset{a_0}{\max} \left\{ IG{_0^0}^{in} + \expec{\ZSp_1} \left[ J_1^{\star} \right] \right\},
	\end{equation}
	and, in general, for each $t \in \left[ 1,T-1 \right]$
	\begin{equation}
		J_t^{\star} = \underset{a_t}{\max} \left\{ \igseqinv + \expec{\ZSp_{t+1}} \left[ J_{t+1}^{\star} \right] \right\}.
	\end{equation}
	We observe that this is the Bellman optimality equation with a new reward, $\rho_t^{in} \triangleq \igseqinv$. This eventually means that Solving the $\rho$-POMDP optimization problem with this reward is equivalent to solving it with the original reward we have started with, $\rho_t = \igseq$.

\begin{figure}
	\centering
	\begin{subfigure}[b]{0.49\textwidth}
		\centering
		\scalebox{0.25}{
			\includegraphics{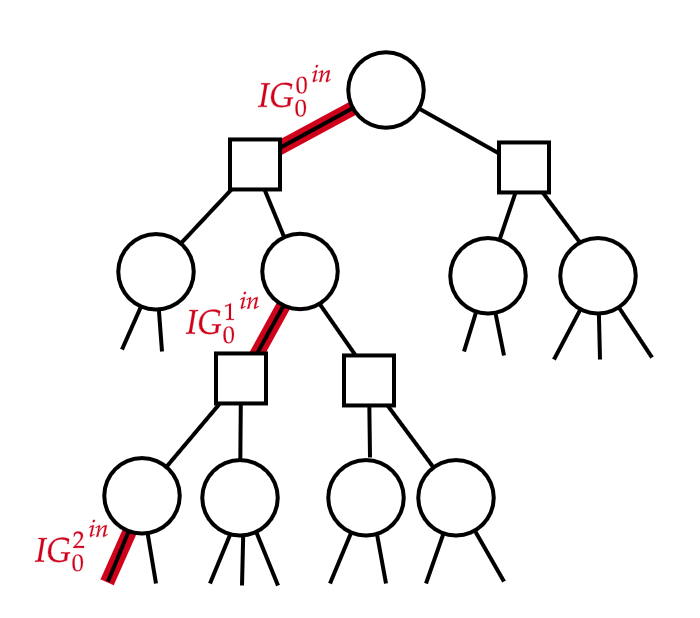}
		}
		\caption{}
		\label{fig:involved_belief_tree_ig}
	\end{subfigure}
	\hfill
	\begin{subfigure}[b]{0.49\textwidth}
		\centering
		\scalebox{0.25}{
			\includegraphics{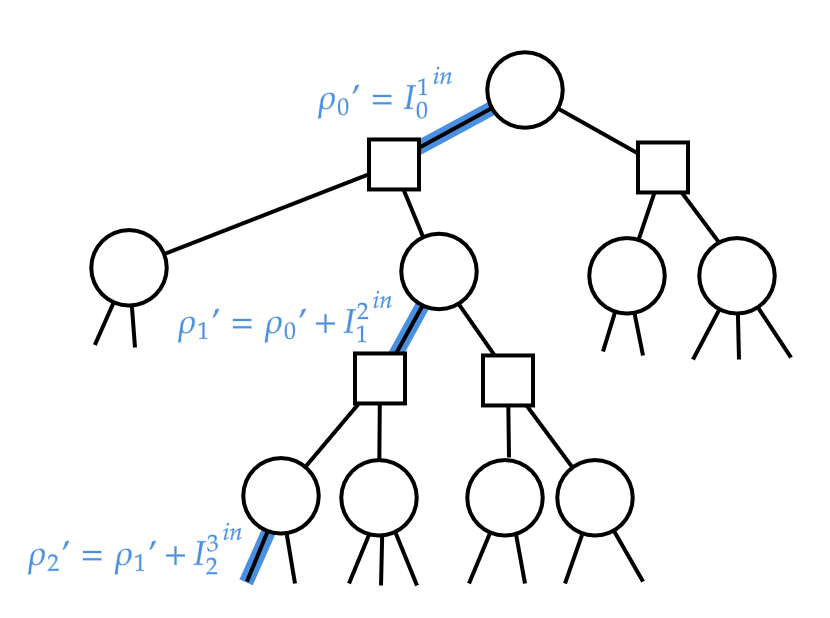}
		}
		\caption{}
		\label{fig:involved_belief_tree_mi}
	\end{subfigure}
	\hfill
	\begin{subfigure}[b]{0.49\textwidth}
		\centering
		\scalebox{0.25}{
			\includegraphics{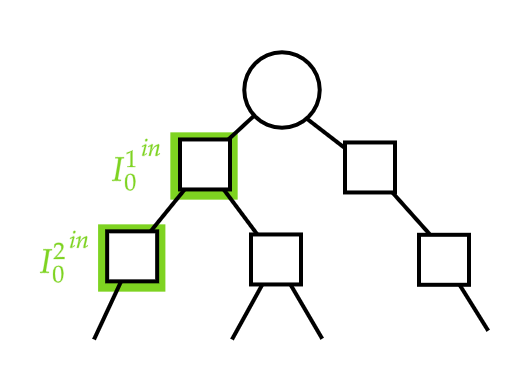}
		}
		\caption{}
		\label{fig:degenerate_belief_tree}
	\end{subfigure}
	
	\caption{Belief trees over the involved variables only, which resemble optimization problems equivalent to the original optimization problem. Circles are observation nodes, squares are action nodes. (a) shows a belief tree where the rewards are {\color{red} sequential involved IGs}; (b) shows a belief tree where the rewards are updated incrementally with {\color{blue} consecutive involved MIs}; (c) shows the resultant degenerate belief tree when trying to directly go through the calculation of the {\color{green} sequential involved MIs}. It is degenerate in the sense that there are only action nodes, without observation nodes.}
	\label{fig:involved_belief_trees}
\end{figure}

This, in turn, means that any optimization solver suitable for the original problem, with the reward $\rho_t$, is also suitable when changing it to $\rho_t^{in}$. This is a key result, since in general these rewards are not equal, however it is much more efficient to calculate $\rho_t^{in}$, as already discussed for the MI case. The belief tree which resembles this new equivalent optimization problem is shown in Figure~\ref{fig:involved_belief_tree_ig}. We can then use the IG definition and calculate it through the entropy terms. This result is again general, but the non-parametric setting then necessitate the usage of entropy estimators, plenty of which exist in the literature, as already discussed.

\begin{prop}
	\label{pr:degenerate-tree}
	Naively calculating the values $\miseq$ yields a degenerate belief tree, in which there are only action nodes, without observation nodes.
\end{prop}


	\subsection*{Proof}
	We remind eq.~(\ref{eq:objective_mi_involved}) is
	\begin{equation}
		J_0^{\star} = \underset{a_{0:T-1}}{\max} \left\{ \sum_{t=0}^{T} \miseqinv \right\}.
	\end{equation}
	Since by definition $I{_0^0}^{in}=0$, we can start the summation from $t=1$
	\begin{equation}
		J_0^{\star} = \underset{a_{0:T-1}}{\max} \left\{ \sum_{t=1}^T \miseqinv \right\}.
	\end{equation}
	We then separate the first action $a_0$ from the rest of the actions $a_{1:T-1}$. We also observe that $I{_0^1}^{in}$ is not a function of $a_{1:T-1}$. This yields
	\begin{equation}
		J_0^{\star} = \underset{a_0}{\max} \left\{ I{_0^1}^{in} + \underset{a_{1:T-1}}{\max} \left\{ \sum_{t=2}^{T-1} \miseqinv \right\} \right\}.
	\end{equation}
	We then observe that the term $\underset{a_{1:T-1}}{\max} \left\{ \sum_{t=2}^{T-1} \miseqinv \right\}$ is equal to $J_1^{\star}$, which yields the following recursive form
	\begin{equation}
		J_0^{\star} = \underset{a_0}{\max} \left\{ I{_0^1}^{in} + J_1^{\star} \right\},
	\end{equation}
	and, in general, $\forall t \in \left[ 1,T-1 \right]$
	\begin{equation}
		J_t^{\star} = \underset{a_t}{\max} \left\{ I{_0^{t+1}}^{in} + J_{t+1}^{\star} \right\}.
	\end{equation}
	We observe that this recursive form is slightly different than the Bellman optimality equation. The Bellman optimality equation, as can be seen in eq.~(\ref{eq:bellman_optimality}), includes also expectation over future observations, while in this formulation it is omitted (more specifically, it is considered at the level of calculating the values $\miseq$). This, in turn, means that a corresponding tree will lack observation nodes, thus it will be a degenerate belief-tree. We note again that this is the result of naively going through direct calculations of the values $\miseq$.


%
This type of a tree can be seen in Figure~\ref{fig:degenerate_belief_tree}. We do not here analyze whether this formulation is good or bad compared to the standard formulation nor whether it would even suit a policy formulation or not. We leave it for future research. We cling to the fact that none of the state-of-the-art tree-based solvers work this way, and suggest another approach.
\begin{lem}
	\label{le:seq-con-mi}
	Let \( \miseq \) denote a sequential MI between times \(0\) and \(t\),
	and \( \micon{i} = \miaug{\X_{i-1}}{x_i}{\Z_i \mid h_i^{-}}\) denote a consecutive MI between times \(i-1\) and \(i\),
	where \(\his_i^{-} = \left\{ z_{1:i-1}, a_{0:i-1} \right\}\) is the history up to time \(i\), without the last observation \(z_i\).
	The sequential MI can be decomposed into multiple consecutive MI values, such that
	\begin{equation}
		\miseq \bydef \sum_{i=1}^t \bigg[ \expec{\ZSp_{1:i-1}} \Big[ \micon{i} \Big] \bigg].
	\end{equation}
\end{lem}


	\subsection*{Proof}
	We remind that the sequential augmented MI is defined as
	\begin{equation}
		\miseq{} \bydef
		\miaug{\X_0}{x_{1:t}}{\Z_{1:t} \mid a_{0:t-1}}
		\bydef
		\expec{\mathcal{Z}_{1:t}} \Big[ \igaug{\X_0}{x_{1:t}}{\Z_{1:t}=\z_{1:t} \mid a_{0:t-1}} \Big].
	\end{equation}
	Detaching the observations $\Z_{l+1:t}$, where $0<l<t$, and expressing the augmented IG with entropies, we get
	\begin{equation}
		\miaug{\X_0}{x_{1:t}}{\Z_{1:t} \mid a_{0:t-1}}
		=
		\expec{\mathcal{Z}_{1:l}} \bigg[ \expec{\mathcal{Z}_{l+1:t}} \Big[ \ent{\X_0} - \ent{\X_t \mid \his_t} \Big] \bigg].
	\end{equation}
	Adding and subtracting the term $\ent{\X_{l} \mid \his_{l}}$, it becomes
	\begin{equation}
		\begin{aligned}
			\miaug{\X_0}{x_{1:t}}{\Z_{1:t} \mid a_{0:t-1}}
			=
			&\expec{\mathcal{Z}_{1:l}} \bigg[ \expec{\mathcal{Z}_{l+1:t}} \Big[
			\big\{ \ent{\X_0} - \ent{\X_{l} \mid \his_{l}} \big\} + \\
			&+ \big\{ \ent{\X_{l} \mid \his_{l}} - \ent{\X_t \mid \his_t} \big\} \Big] \bigg].
		\end{aligned}
	\end{equation}
	Observing that both new differences are augmented IGs as well, and that the first difference is not a function of the last observation, we get
	\begin{equation}
		\begin{aligned}
			\miaug{\X_0}{x_{1:t}}{\Z_{1:t} \mid a_{0:t-1}}
			=
			&\expec{\mathcal{Z}_{1:l}} \bigg[ \igaug{\X_0}{x_{1:l}}{\z_{1:l} \mid a_{0:l-1}} + \\
			&+ \expec{\mathcal{Z}_{l+1:t}} \Big[
			\igaug{\X_{l}}{x_{l+1:t}}{\z_{l+1:t} \mid a_{0:t-1}, \z_{1:l}} \Big] \bigg],
		\end{aligned}
	\end{equation}
	%
	The expectation over the augmented IG is the augmented MI, and so we get the following recursive form
	\begin{equation}
		\begin{aligned}
			\miaug{\X_0}{x_{1:t}}{\Z_{1:t} \mid a_{0:t-1}}
			=
			& \miaug{\X_{0}}{x_{1:l}}{\Z_{1:l} \mid a_{0:l-1}} + \\
			+ &\expec{\mathcal{Z}_{1:l}} \Big[ \miaug{\X_{l}}{x_{l+1:t}}{\Z_{l+1:t} \mid a_{0:t-1}, \z_{1:l}} \Big].
		\end{aligned}
	\end{equation}
	The specific case of choosing $l = t-1$ yields
	\begin{equation}
		\begin{aligned}
			\miaug{\X_0}{x_{1:t}}{\Z_{1:t} \mid a_{0:t-1}}
			=
			& \miaug{\X_{0}}{x_{1:t-1}}{\Z_{1:t-1} \mid a_{0:t-2}} + \\
			&+ \expec{\mathcal{Z}_{1:t-1}} \Big[ \miaug{\X_{t-1}}{x_t}{\Z_t \mid h_t^{-}} \Big],
		\end{aligned}
		\label{eq:recursive}
	\end{equation}
	where $\his_t^{-} = \left\{ z_{1:t-1}, a_{0:t-1} \right\}$ is the history up to time $t$, without the last observation $z_t$.
	Opening the recursive form of the sequential augmented MI in eq.~(\ref{eq:recursive}) yields
	\begin{equation}
		\begin{gathered}
			\miaug{\X_0}{x_{1:t}}{\Z_{1:t} \mid a_{0:t-1}}
			=
			\miaug{\X_0}{x_1}{\Z_1 \mid h_1^{-}} + \\
			+ \expec{\mathcal{Z}_{1}} \Big[ \miaug{\X_1}{x_2}{\Z_2 \mid \his_2^{-}} \Big] + \dots
			+ \expec{\mathcal{Z}_{1:t-1}} \Big[ \miaug{\X_{t-1}}{x_t}{\Z_t \mid \his_t^{-}} \Big],
		\end{gathered}
	\end{equation}
	which can more compactly be written as
	\begin{equation}
		\miaug{\X_0}{x_{1:t}}{\Z_{1:t} \mid a_{0:t-1}} =
		\sum_{i=1}^t
		\bigg[ \expec{\ZSp_{1:i-1}} \Big[
		\miaug{\X_{i-1}}{x_i}{\Z_i \mid \his_i^{-}}
		\Big] \bigg].
	\end{equation}
	Returning to the short notations, we finally get
	\begin{equation}
		\miseq = \sum_{i=1}^t \bigg[ \expec{\ZSp_{1:i-1}}
		\Big[ \micon{i} \Big] \bigg].
	\end{equation}


%
The main result of Theorem 1 can be applied on both the sequential and the consecutive MI values by assigning the notations in a slightly different manner, such that the result of Lemma~\ref{le:seq-con-mi} is transformed into
\begin{equation}
	\label{eq:seq-con-mi-inv}
	\miseqinv = \sum_{i=1}^{t} \left[ \expec{\ZSp_{1:i-1}} \left[ \miconinv{i} \right] \right],
\end{equation}
where $\miconinv{i} = \miaug{\X_{i-1}^{in}}{x_i}{\Z_i \mid h_i^{-}}$ is the consecutive MI over the involved subset of the state $\X_{i-1}$.

\begin{theorem}
	\label{th:involved-expected-reward}
	Let us define a new reward, \( \rho_t' = \sum_{i=1}^{t+1} \miconinv{i} \). Solving the \( \rho \)-POMDP optimization problem with this reward is equivalent to solving it with the original reward, \( \rho_t = \igseq \),
	such that
	\begin{equation}
		J_t^{\star} = \underset{a_t}{\max} \left\{ \rho_t' + \expec{\ZSp_{t+1}} \left[ J_{t+1}^{\star} \right] \right\}.
	\end{equation}
\end{theorem}


	\subsection*{Proof}
	We remind eq.~(\ref{eq:objective_mi_involved}) is
	\begin{equation}
		J_0^{\star} = \underset{a_{0:T-1}}{\max} \left\{ \sum_{t=0}^{T} \miseqinv \right\}.
	\end{equation}
	Since by definition $I{_0^0}^{in}=0$, we can start the summation from $t=1$
	\begin{equation}
		J_0^{\star} = \underset{a_{0:T-1}}{\max} \left\{ \sum_{t=1}^T \miseqinv \right\}.
	\end{equation}
	Plugging the result from eq.~(\ref{eq:seq-con-mi-inv}) into the above yields
	\begin{equation}
		J_0^{\star} = \underset{a_{0:T-1}}{\max} \left\{ \sum_{t=1}^T \left[ \sum_{i=1}^t \left[ \expec{\ZSp_{1:i-1}} \left[ \miconinv{i} \right] \right] \right] \right\}.
	\end{equation}
	Due to commutativity, we can switch between the order of expectation and summation, which yields
	\begin{equation}
		J_0^{\star} = \underset{a_{0:T-1}}{\max} \left\{ \expec{\ZSp_{1:T-1}} \left[ \sum_{t=1}^T \left[ \sum_{i=1}^t \left[ \miconinv{i} \right] \right] \right] \right\}.
	\end{equation}
	We then denote $\rho_{t-1}' \bydef \sum_{i=1}^t \left[ \miconinv{i} \right]$, and get
	\begin{equation}
		J_0^{\star} = \underset{a_{0:T-1}}{\max} \left\{ \expec{\ZSp_{1:T-1}} \left[ \sum_{t=1}^T \rho_{t-1}' \right] \right\}
		= \underset{a_{0:T-1}}{\max} \left\{ \expec{\ZSp_{1:T-1}} \left[ \sum_{t=0}^{T-1} \rho_t' \right] \right\}.
	\end{equation}
	We then separate the first action $a_0$ from the rest of the actions $a_{1:T-1}$. We also observe that $\Z_1$ is not a function of $a_{1:T-1}$, and that $\rho_0' = I{_0^1}^{in}$ is not a function of both $a_{1:T-1}$ and $\Z_1$ (since $I_0^1$ is already an expectation over $\Z_1$). This yields
	\begin{equation}
		J_0^{\star} = \underset{a_0}{\max} \left\{ \rho_0' + \expec{\ZSp_1} \left[ \underset{a_{1:T-1}}{\max} \left\{ \expec{\ZSp_{2:T-1}} \left[ \sum_{t=1}^{T-1} \rho_t' \right] \right\} \right] \right\}.
	\end{equation}
	We then observe that the term inside the expectation over $\Z_1$ is equal to $J_1^{\star}$, which yields the following recursive form
	\begin{equation}
		J_0^{\star} = \underset{a_0}{\max} \left\{ \rho_0' + \expec{\ZSp_1} \left[ J_1^{\star} \right] \right\},
	\end{equation}
	and, in general, $\forall t \in \left[ 1,T-1 \right]$
	\begin{equation}
		J_t^{\star} = \underset{a_t}{\max} \left\{ \rho_t' + \expec{\ZSp_{t+1}} \left[ J_{t+1}^{\star} \right] \right\}.
	\end{equation}
	We observe that this is the Bellman optimality equation with the new reward, $\rho_t'$. This eventually means that Solving the $\rho$-POMDP optimization problem with this reward is equivalent to solving it with the original reward we have started with, $\rho_t = \igseq$. We note another slight difference between the formulations, for which the latter formulation does not include a terminal reward.


This allows the usage of estimators which directly estimate MI, as our suggested estimator \mismc{} does, together with the usage of tree-based solvers of $\rho$-POMDP. However, we emphasize that instead of sequential MI values, we will calculate consecutive MI values.

We note that $\miconinv{i} = \underset{\mathcal{Z}_{i}}{\mathbb{E}}\left[ IG_{i-1}^{i}\right]$. This means that the calculation of the MI values is not limited only to the observations that are used for constructing the tree, thus the calculation can be more accurate, which is another added value of this formulation.

And, lastly, we note that $\rho_t' = \sum_{i=1}^{t+1} \left[ \miconinv{i} \right] = \rho_{t-1}' + I{_{t}^{t+1}}^{in}$. Meaning that for each node, we can calculate the reward based on the previous reward and just update the new information incrementally, without having to calculate the entire reward from scratch. The belief tree which resembles this optimization problem is shown in Figure~\ref{fig:involved_belief_tree_mi}.

Using one-time marginalization, i.e. determining ahead all the involved variables (together with variables which are required for other reward functions), and marginalizing out the rest of the variables, the above analysis suggests that the entire tree can be constructed considering only the marginalized beliefs rather than the entire-state beliefs. This, in turn, reduces also the complexity of constructing this tree, since we avoid maintaining and propagating the beliefs over unnecessary states.
Care should be taken, however, when using this approach, since marginalizing out a variable which would in retrospect be found to be involved would mean that the tree should be updated from the root. Also note that this approach might prevent the usage of calculation re-use approaches (e.g. \cite{Farhi17icra}, \cite{Farhi19icra}) since we only consider a subset of the state for the whole planning process.


%

%% file: paper.bbl
\begin{thebibliography}{10}
	
	\bibitem{Beirlant97jmss}
	Jan Beirlant, Edward~J Dudewicz, L{\'a}szl{\'o} Gy{\"o}rfi, Edward~C Van~der
	Meulen, et~al.
	\newblock Nonparametric entropy estimation: An overview.
	\newblock {\em International Journal of Mathematical and Statistical Sciences},
	6(1):17--39, 1997.
	
	\bibitem{Boers10fusion}
	Y.~{Boers}, H.~{Driessen}, A.~{Bagchi}, and P.~{Mandal}.
	\newblock Particle filter based entropy.
	\newblock In {\em 2010 13th International Conference on Information Fusion},
	pages 1--8, 2010.
	
	\bibitem{Chli09thesis}
	Margarita Chli.
	\newblock {\em Applying Information Theory to Efficient {SLAM}}.
	\newblock PhD thesis, Imperial College London, 2009.
	
	\bibitem{Elimelech17icra}
	Khen Elimelech and Vadim Indelman.
	\newblock Consistent sparsification for efficient decision making under
	uncertainty in high dimensional state spaces.
	\newblock In {\em IEEE Intl. Conf. on Robotics and Automation (ICRA)}, pages
	3786--3791, 05 2017.
	
	\bibitem{Farhi17icra}
	E.~I. Farhi and V.~Indelman.
	\newblock Towards efficient inference update through planning via jip - joint
	inference and belief space planning.
	\newblock In {\em IEEE Intl. Conf. on Robotics and Automation (ICRA)}, 2017.
	
	\bibitem{Farhi19icra}
	E.~I. Farhi and V.~Indelman.
	\newblock ix-bsp: Belief space planning through incremental expectation.
	\newblock In {\em IEEE Intl. Conf. on Robotics and Automation (ICRA)}, May
	2019.
	
	\bibitem{Fischer20icml}
	Johannes Fischer and Omer~Sahin Tas.
	\newblock Information particle filter tree: An online algorithm for pomdps with
	belief-based rewards on continuous domains.
	\newblock In {\em Intl. Conf. on Machine Learning (ICML)}, Vienna, Austria,
	2020.
	
	\bibitem{Huang21icra}
	Qiangqiang Huang, Can Pu, Dehann Fourie, Kasra Khosoussi, Jonathan~P How, and
	John~J Leonard.
	\newblock Nf-isam: Incremental smoothing and mapping via normalizing flows.
	\newblock In {\em IEEE Intl. Conf. on Robotics and Automation (ICRA)}, 2021.
	
	\bibitem{Indelman16ral}
	V.~Indelman.
	\newblock No correlations involved: Decision making under uncertainty in a
	conservative sparse information space.
	\newblock {\em IEEE Robotics and Automation Letters (RA-L)}, 1(1):407--414,
	2016.
	
	\bibitem{Kopitkov17ijrr}
	D.~Kopitkov and V.~Indelman.
	\newblock No belief propagation required: Belief space planning in
	high-dimensional state spaces via factor graphs, matrix determinant lemma and
	re-use of calculation.
	\newblock {\em Intl. J. of Robotics Research}, 36(10):1088--1130, August 2017.
	
	\bibitem{Kurniawati08rss}
	H.~Kurniawati, D.~Hsu, and W.~S. Lee.
	\newblock {SARSOP}: Efficient point-based {POMDP} planning by approximating
	optimally reachable belief spaces.
	\newblock In {\em Robotics: Science and Systems (RSS)}, 2008.
	
	\bibitem{Kurniawati16chapter}
	Hanna Kurniawati and Vinay Yadav.
	\newblock An online pomdp solver for uncertainty planning in dynamic
	environment.
	\newblock In {\em Robotics Research}, pages 611--629. Springer, 2016.
	
	\bibitem{Platt11isrr}
	R.~Platt, L.~Kaelbling, T.~Lozano-Perez, and R.~Tedrake.
	\newblock Efficient planning in non-gaussian belief spaces and its application
	to robot grasping.
	\newblock In {\em Proc. of the Intl. Symp. of Robotics Research (ISRR)}, 2011.
	
	\bibitem{Ryan08gnc}
	Allison Ryan.
	\newblock Information-theoretic tracking control based on particle filter
	estimate.
	\newblock In {\em AIAA Guidance, Navigation and Control Conference}, pages
	1--15, 2008.
	
	\bibitem{Silver10nips}
	David Silver and Joel Veness.
	\newblock Monte-carlo planning in large pomdps.
	\newblock In {\em Advances in Neural Information Processing Systems (NIPS)},
	pages 2164--2172, 2010.
	
	\bibitem{Stachniss05rss}
	C.~Stachniss, G.~Grisetti, and W.~Burgard.
	\newblock Information gain-based exploration using {Rao-Blackwellized} particle
	filters.
	\newblock In {\em Robotics: Science and Systems (RSS)}, pages 65--72, 2005.
	
	\bibitem{Stowell09spl}
	Dan Stowell and Mark~D Plumbley.
	\newblock Fast multidimensional entropy estimation by $k$-d partitioning.
	\newblock {\em IEEE Signal Processing Letters}, 16(6):537--540, 2009.
	
	\bibitem{Sunberg18icaps}
	Zachary Sunberg and Mykel Kochenderfer.
	\newblock Online algorithms for pomdps with continuous state, action, and
	observation spaces.
	\newblock In {\em Proceedings of the International Conference on Automated
		Planning and Scheduling}, volume~28, 2018.
	
	\bibitem{Ye17jair}
	Nan Ye, Adhiraj Somani, David Hsu, and Wee~Sun Lee.
	\newblock Despot: Online pomdp planning with regularization.
	\newblock {\em JAIR}, 58:231--266, 2017.
	
	\bibitem{Zhang20ijrr}
	Zhengdong Zhang, Theia Henderson, Sertac Karaman, and Vivienne Sze.
	\newblock Fsmi: Fast computation of shannon mutual information for
	information-theoretic mapping.
	\newblock {\em Intl. J. of Robotics Research}, 39(9):1155--1177, 2020.
	
\end{thebibliography}
